\documentclass[10pt, a4paper]{article}

\usepackage[final]{lrec2026} 

\usepackage{booktabs}
\usepackage{amsmath}
\usepackage{graphicx}
\usepackage{subcaption} 

\usepackage{float} 

\title{Are the LLMs Capable of Maintaining at Least the Language Genus?}

\name{Sandra Mitrović$^{*, 1}$\thanks{* Equal contribution.}, David Kletz$^{*, 1}$\footnotemark[1], Ljiljana Dolamic$^2$, Fabio Rinaldi$^{1}$} 

\address{$^{1}$ SUPSI, IDSIA, Switzerland  \\
          $^{2}$armasuisse, Science \& Technology, Switzerland\\
         \{sandra.mitrovic, david.kletz, fabio.rinaldi\}@supsi.ch\\ 
         ljiljana.dolamic@armasuisse.ch\\}

\abstract{
Large Language Models (LLMs) display notable variation in multilingual behavior, yet the role of genealogical language structure in shaping this variation remains underexplored. In this paper, we investigate whether LLMs exhibit sensitivity to linguistic genera by extending prior analyses on the MultiQ dataset. We first check if models prefer to switch to genealogically related languages when prompt language fidelity is not maintained. Next, we investigate whether knowledge consistency is better preserved within than across genera. We show that genus-level effects are present but strongly conditioned by training resource availability. We further observe distinct multilingual strategies across LLMs families. Our findings suggest that LLMs encode aspects of genus-level structure, but training data imbalances remain the primary factor shaping their multilingual performance.
 \\ \newline }

\begin{document}

\maketitleabstract

\section{Introduction}\label{s:intro}

 
 Numerous studies have investigated how variations in the input prompt affect the outputs of Large Language Models (LLMs) \citep{habba-etal-2025-dove, liu2025selfreflectionmakeslargelanguage, zhou-etal-2023-survival}. Even superficial modifications that leave the semantic content unchanged can yield substantial differences in model responses --for example, altering the order of proposed answers in multiple-choice benchmarks \citep{alzahrani-etal-2024-benchmarks} or reordering few-shot examples \citep{pmlr-v139-zhao21c}.
A particularly salient dimension of this phenomenon is the language of the prompt. For instance, \citet{bandarkar-etal-2024-belebele} show that gpt3.5-turbo and Llama-2-chat perform dramatically better --by 40.8 and 25.4 points, respectively-- on a QA task when questions are posed in English rather than Icelandic.

Beyond performance differences, research has highlighted a related phenomenon: infidelity to the prompt language. LLMs frequently generate responses in a language different from that of the input \citep{shaham-etal-2024-multilingual, liu2025xragcrosslingualretrievalaugmentedgeneration}. When queried in Arabic, models may partially or fully switch to English \citep{chen-etal-2025-large}. Even when posed mathematical questions in non-English languages, language-aligned LLMs often produce English chain-of-thought reasoning before providing the final answer \citep{tran2025scalingtesttimecomputelowresource, zhu-etal-2024-question}. Several studies have documented systematic patterns of such language-switching behaviors across different models and tasks \citep{wisniewski-etal-2025-exploring, almasi-kristensen-mclachlan-2025-alignment}.

While existing research has primarily examined language-switching as a binary phenomenon (adherence vs. deviation from the prompt language), the linguistic structure underlying these behaviors remains underexplored.
In this paper, we introduce a genealogical perspective on multilingual LLM behavior, investigating whether linguistic proximity --as defined by genealogical classification-- correlates with consistency in model outputs.
Our central hypothesis is that LLMs may encode a form of genus-level coherence, potentially leading to more stable behaviors within linguistic families than across them.


\paragraph{Research questions}

We explore this potential coherence through two complementary analyses:

\begin{enumerate}
    \item Genus fidelity: When LLMs fail to respond in the prompt language, do they preferentially switch to another language of the same genus?

    \item Knowledge sharing across a genus: If an LLM answers a question correctly in one language, is it also likely to answer correctly when the same question is asked in another language of the same genus?

\end{enumerate}

    Throughout the paper, we refer to the prompt’s original language as the source language, its translation as the target language, and the LLM’s response language as the generation language\footnote{Our code is available on github at 
    \url{https://github.com/IDSIA-NLP/GenusPref}
    }.

\section{Methodology and data}\label{s:method}

Our methodology builds primarily on the work of \citetlanguageresource{holtermann-etal-2024-evaluating}, who introduced the MultiQ dataset for evaluating multilingual capabilities of LLMs. Rather than conducting new large-scale data collection, we leverage these existing resources to perform a targeted secondary analysis focused on genealogical effects --a dimension not explored in the original study. This approach allows us to benefit from MultiQ's extensive coverage and rigorous design while introducing our novel genealogical perspective on multilingual LLM behavior.

\subsection{The MultiQ dataset \citep{holtermann-etal-2024-evaluating}}

\citet{holtermann-etal-2024-evaluating} developed MultiQ to investigate fundamental multilingual capabilities of LLMs through a large-scale parallel question-answering dataset comprising 27,400 questions across 137 languages. The dataset covers diverse question types (open-ended,  closed-ended, reasoning questions) and domains (chemistry, physics, astronomy, history, maths, geography, art, sports, music, animals).
The original study evaluated LLMs along two primary dimensions:

\textit{Language fidelity}: Whether the model generates its response in the same language as the input prompt.

\textit{Question-answering accuracy}: Whether the generated response is factually correct.

Their findings revealed substantial variation both across models and across languages, highlighting critical gaps in multilingual alignment. However, their language grouping strategy --while methodologically sound for their research questions-- was too coarse for our genealogical analysis.

\paragraph{Model Selection}
To ensure direct comparability with existing results, we analyze the same four models evaluated by \citet{holtermann-etal-2024-evaluating}: Llama-2-7B-Chat \citep{touvron2023llama2openfoundation},  Mistral-7B-Instruct-v0.1 \citep{jiang2023mistral7b}, Mixtral-8x7B-Instruct-v0.1 \citep{jiang2024mixtralexperts}, Qwen1.5-7B-Chat \citep{bai2023qwentechnicalreport}

\paragraph{Apertus-8B}

The models evaluated in MultiQ are Open-Weight Models, but not Fully Open Models \citep{swissai2025apertus}, as their training data are not completely transparent. However, we consider it essential to include in our results at least one Fully Open Model.
We therefore select Apertus-8B, a multilingual model whose pre-training data are fully disclosed \citep{swissai2025apertus}. According to its authors, Apertus was trained on data covering approximately 1,800 languages, with around 40\% of the corpus being non-English.

To integrate Apertus into our evaluation, we generate responses to the 200 questions in 137 languages of the MultiQ benchmark, and assess them both for answer correctness and language identification.
To ensure full comparability, we strictly replicate the configuration of \citet{holtermann-etal-2024-evaluating}: we run their publicly released code\footnote{Available at \url{https://github.com/paul-rottger/multiq/tree/main}} and use the same evaluation models, namely \texttt{gpt-4-0125-preview} \citep{openai2024gpt4technicalreport} for answer quality assessment and \texttt{cis-lmu/glotlid, model\_v2}\footnote{\url{https://huggingface.co/cis-lmu/glotlid}} \citep{kargaran2023glotlid} for language detection.

\subsection{From Family-Level to Genus-Level Classification}

The original MultiQ analysis grouped languages into three broad categories: English, Same (languages from the same family as the source language), and Other. 
We argue that for our RQs  this classification proves insufficient as language families are too broad and mix together languages of different characteristics.

Consider, for example, the Indo-European family: it encompasses both English --which dominates LLM training corpora \citep{zhong2024englishcentricllmslanguagemultilingual, csaki-etal-2024-sambalingo, gupta2025multilingualperformancebiaseslarge} 
-- and Hindi, which 
remains underrepresented in training data. These two languages are very different in many aspects. With respect to syntax, English heavily relies on subject-verb-object order, whereas Hindi uses subject-object-verb order; with respect to the writing system, English uses the Latin alphabet and Hindi uses Devanagari; with respect to vocabulary, English mostly borrows from Latin while Hindi has borrowings from Persian and Arabic~\citep{masica1993indo, shapiro1989primer}. Therefore, as these languages differ substantially in their representation within LLM training corpora and their structural characteristics, making family-level grouping is potentially misleading.

To address this limitation, we adopt a more fine-grained, genus-level classification, which provides optimal granularity for our analysis. Genus represents an intermediate taxonomic level in linguistic classification --more specific than family but broader than individual languages-- making it well-suited for detecting systematic patterns while maintaining sufficient statistical power.

\paragraph{Language Coding and Genus Mapping}

Our genus-level analysis required careful alignment given that multiple coding systems were used in MultiQ. More specifically:
\begin{itemize}
    \item Source languages are annotated with Google Translate IDs and WALS codes
    \item Generation languages are automatically identified using GlotLID \citep{kargaran2023glotlid}, which assigns ISO 639-3 codes
\end{itemize}

We map all languages to their corresponding genera using the World Atlas of Language Structures (WALS) database \citet{wals}, which contains 2,662 language entries, each annotated with genealogical information including genus classification. WALS provides both WALS-specific codes and ISO 639-3 codes, enabling consistent cross-referencing across the different identifier systems used in MultiQ.
This mapping process involved: 1) a direct mapping: Languages with existing WALS codes were directly mapped to their genera and 2) ISO code alignment: Languages identified only by ISO 639-3 codes were matched to WALS entries.

The resulting genus mapping covers 47 genera across 21 language families, providing sufficient diversity for robust statistical analysis while maintaining genealogical precision.

Our analysis extends the original MultiQ evaluation framework in two key dimensions:
\begin{enumerate}
    \item Genus Fidelity Analysis: We examine whether language-switching patterns respect genealogical boundaries by comparing within-genus vs. cross-genus switching rates;
    \item Cross-Genus Consistency Analysis: We assess whether question-answering accuracy correlates more strongly within genealogical groups than across them.
\end{enumerate}



\section{Genus fidelity}\label{s:fidel}

In this section, we investigate whether LLMs display systematic biases toward or within specific language genera. Our primary focus is on genus-level fidelity, as this provides the most direct test of genealogical effects. 

\subsection{Methodology}

We operationalize genus fidelity as the tendency for models to generate responses within the same genus as the one of the input prompt. For each source genus, we collect all model responses, classify their genera using our WALS-based mapping, and compute generation distributions.

Our primary metric is \textit{cross-genus model fidelity} which captures the proportion of outputs in which the model faithfully maintains the linguistic genus of the input, thereby providing a quantitative measure of cross-genus consistency.

Formally, we define the FidelityScore for a given model as follows:
\begin{equation}
\begin{split}
\text{FidelityScore}_{LLM} &= \frac{N_{\text{faithful}}}{N_{\text{total}}} \\
&= \frac{|\{p: G_p\!\!=\!\!G_{LLM(p)}\}|}{|\{p\}|}
\end{split}
\end{equation}

where $p$ denotes an input prompt, $LLM(p)$ the corresponding model output, $|.|$ the cardinality of a set and $G_x$ the genus associated with the text $x$.
In other words: 

\[
\begin{aligned}
N_{\text{faithful}} & = \text{Number of prompts where the output}\\
& \quad \quad \text{ genus matches the input genus},\\
N_{\text{total}} & = \text{Total number of prompts.}
\end{aligned}
\]

\subsection{Results}

\paragraph{Model-centric Perspective}
Table~\ref{tab:genus_fidelity_score} presents genus fidelity scores across all evaluated models.

\begin{table}[!h]
\centering
\begin{tabular}{ll}
\toprule
Model & Fidelity \\
\midrule
Llama-2-7b  & 17.3 \\
Llama-2-14b & 23.1 \\
Llama-2-70b & 27.9 \\
Mixtral-8x7B  & 73.9 \\
Mistral-7B  & 75.0 \\
Qwen1.5-7B-Chat    & 70.5 \\
Apertus-8B & \textbf{92.3}\\
\bottomrule
\end{tabular}
\caption{Genus fidelity score by model.}
\label{tab:genus_fidelity_score}
\end{table}

\paragraph{}
The results indicate substantial variation in genus fidelity between models, revealing a clear separation into three fidelity-score tiers. Models from the Llama family get the lowest fidelity scores, with Llama-2-7B achieving as only 0.17 genus consistency. While increasing the model size seems to have positive effect on genus fidelity score, even the largest LLama model Llama-2-70B only reaches 0.28 genus fidelity. Mistral-7B, Mixtral-8x7B and Qwen1.5-7B-Chat have remarkably higher fidelity scores ranging from 0.7 to 0.75 of genus fidelity. The highest performing in terms of genus fidelity is Apertus with 0.92. This suggests that some models have developed stronger genealogical coherence in their multilingual representations. However, it is important to note that  fidelity does not guarantee answer correctness.

\begin{figure*}[!t]
        \centering
        \includegraphics[width=\textwidth]{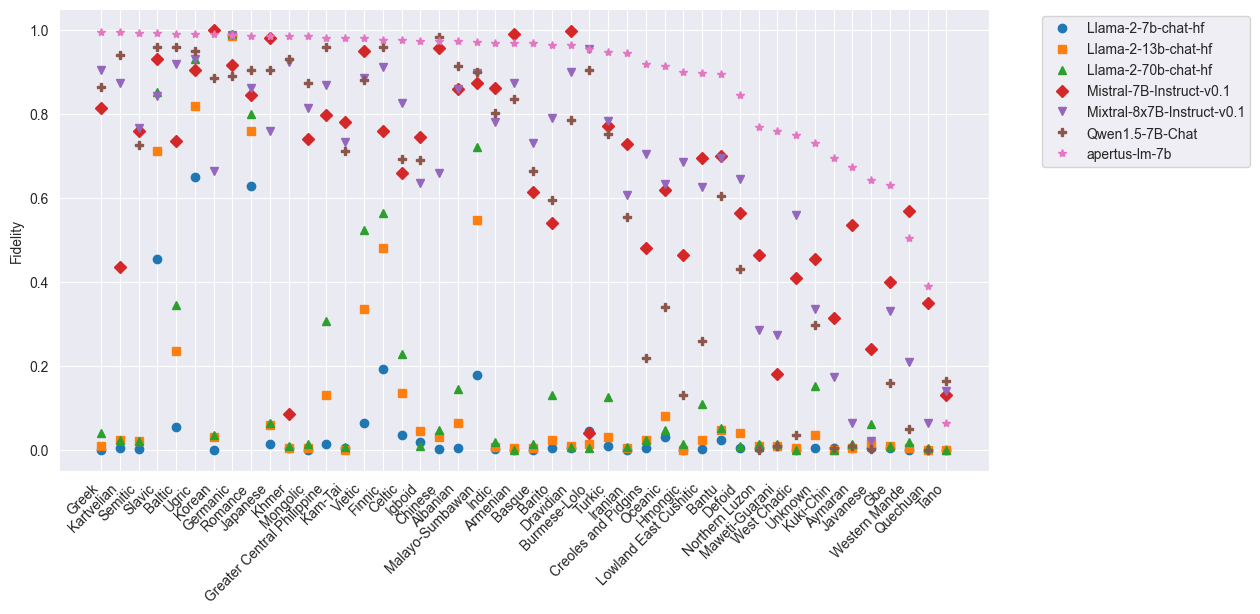}
        \caption{Fidelity scores across genera (existing in MultiQ) and models (existing in MultiQ + Apertus).}
        \label{fig:fidelity_models_all}
\end{figure*}

Next, we take a closer look at each model separately. More specifically, for each model, and every single prompt $p$, we extract the input genus $G_i$ (the genus of the input prompt)
and the output genus $G_o$ (the genus of the model-generated output $LLM(p)$). 
We present the genus fidelity across different genera (present in MultiQ) and models in Figure~\ref{fig:fidelity_models_all}.

The results reveal clear differences in model behaviour across genera. Models from Llama family, although predominantly having low fidelity scores, have extremely high fidelity scores for Germanic genus (around 1,even for the smallest model) and notably high scores for Ugric (larger bigger models above 0.8), Romance and Slavic genera (above 0.7). We also observed a strong tendency of  Llama models  to default to English (a Germanic language) when confronted with non-English prompts.
Mistral, Mixtral and Qwen show a generally high fidelity, though their performances vary accross genera --for example, significantly drops for the Tano, Quechuan and Kuki-Chin genera. Interestingly, Mixtral and Qwen exhibit more siilar behaviour to each other than to Mistral. In particular, for Khmer and Burmese-Lolo genera, Mistral has a fidelity score $\leq 0.15$ while Mixtral and Qwen have score $\geq 0.9$. Similarly, for Vietic, Kam-Tai and Iranian genera, Mixtral and Qwen show closer results. Despite lacking explicit multilingual branding, Mistral often responds in the prompt genus, suggesting robust multilingual competence. Nevertheless, we noticed that on fallbacks these two models behave contrastingly: Mixtral almost always uses English while Qwen is more variable in fallbacks, sometimes producing outputs in unrelated languages (oftentimes in Chinese).
Finally, Apertus demonstrates near-perfect genus fidelity, having score $\leq 0.5$ only for Tano and Quechuan.


Finally, model descriptions do not always correspond to observed behaviour: despite multilingual claim, LLaMA exhibits a systematic English bias, whereas Mistral demonstrates stronger multilingual fidelity despite the absence of explicit multilingual positioning.

\begin{figure*}[ht]
\centering
\begin{subfigure}[b]{0.3\textwidth}
\centering
\includegraphics[width=\textwidth]{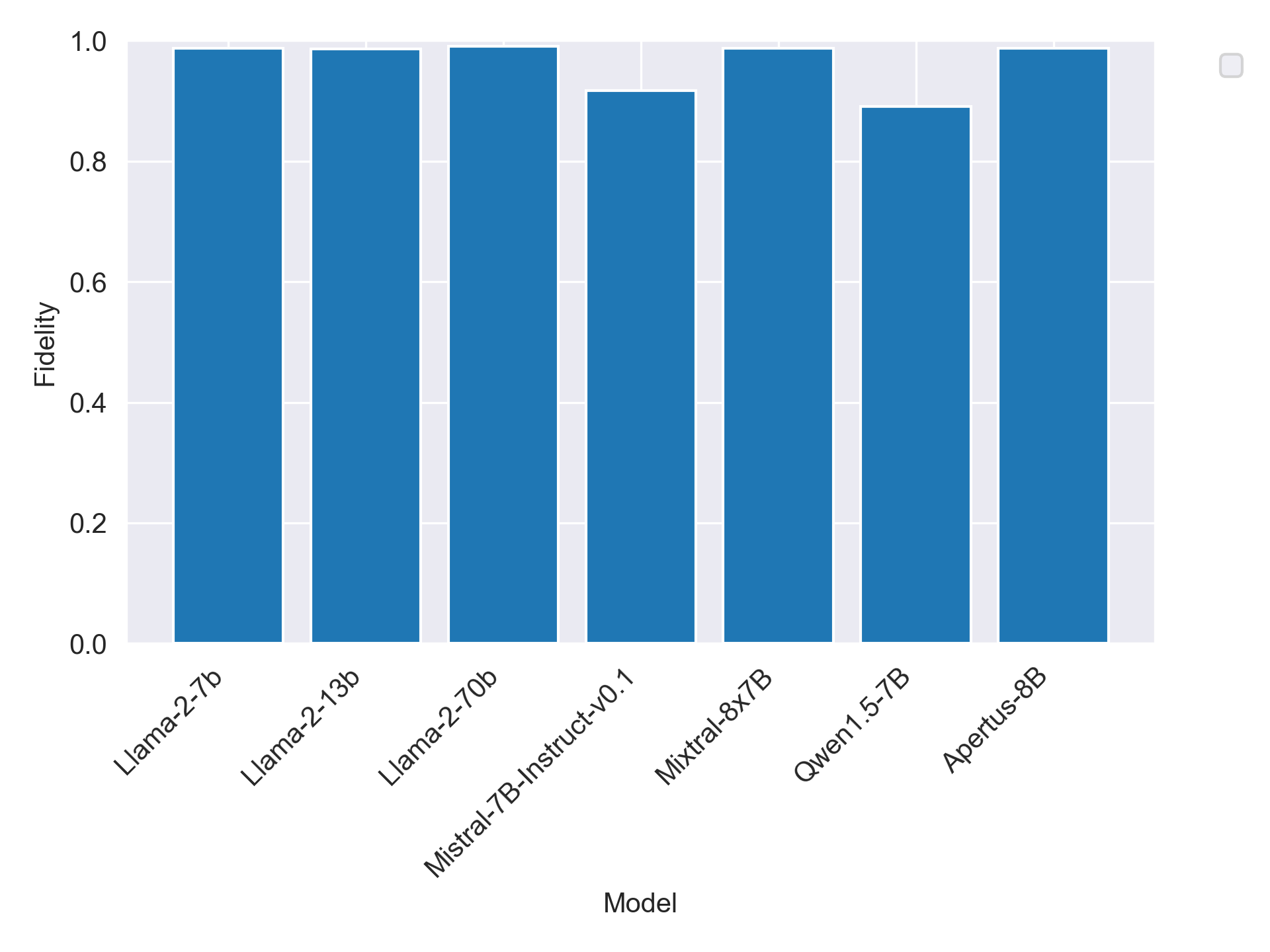}
\caption{Germanic}
\label{fig:fidelity_genus_Germanic}
\end{subfigure}
\hfill
\begin{subfigure}[b]{0.3\textwidth}
    \centering
    \includegraphics[width=\textwidth]{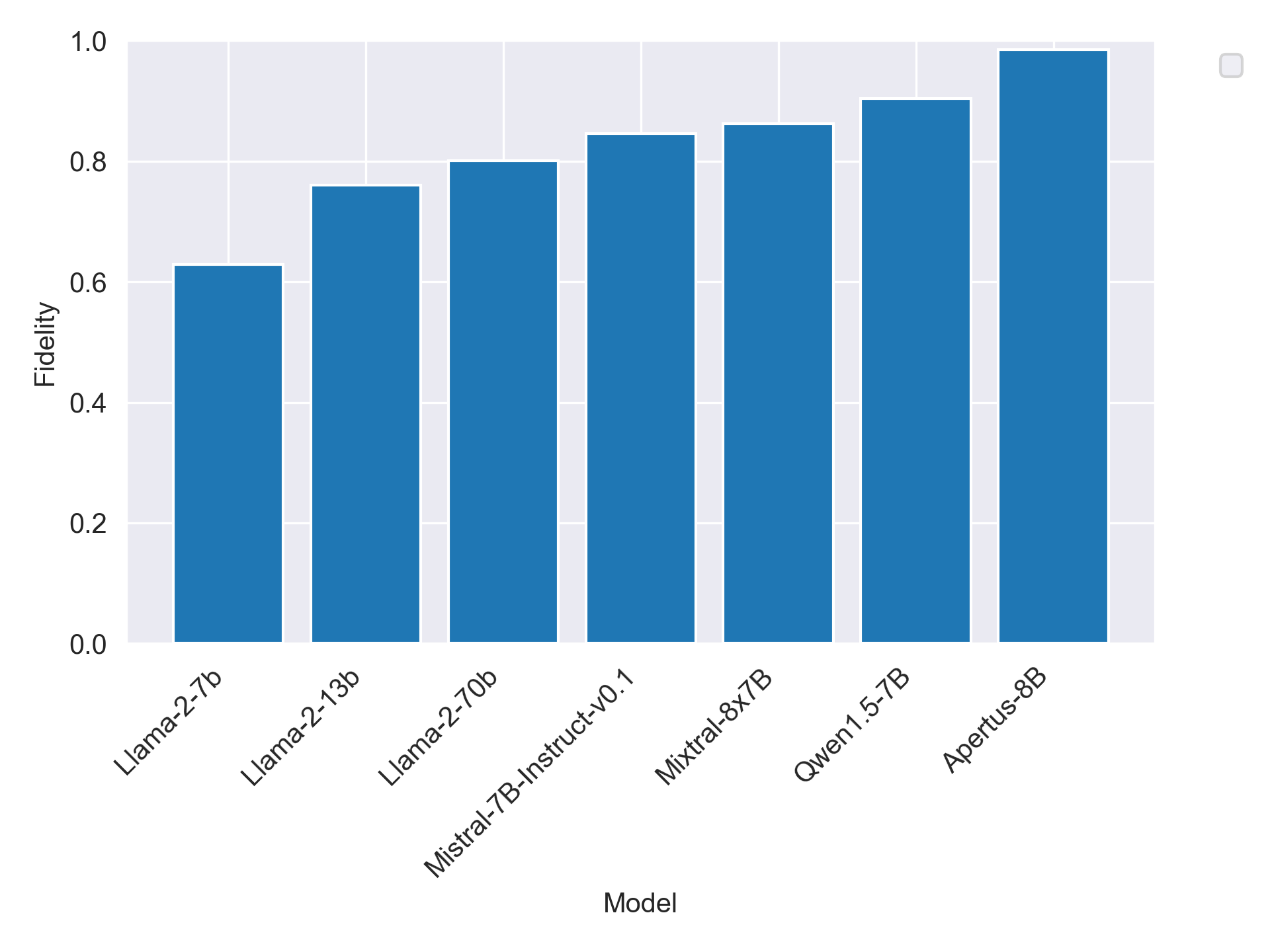}
    \caption{Romance}
    \label{fig:fidelity_genus_Romance}
\end{subfigure}
\hfill
\begin{subfigure}[b]{0.3\textwidth}
    \centering
    \includegraphics[width=\textwidth]{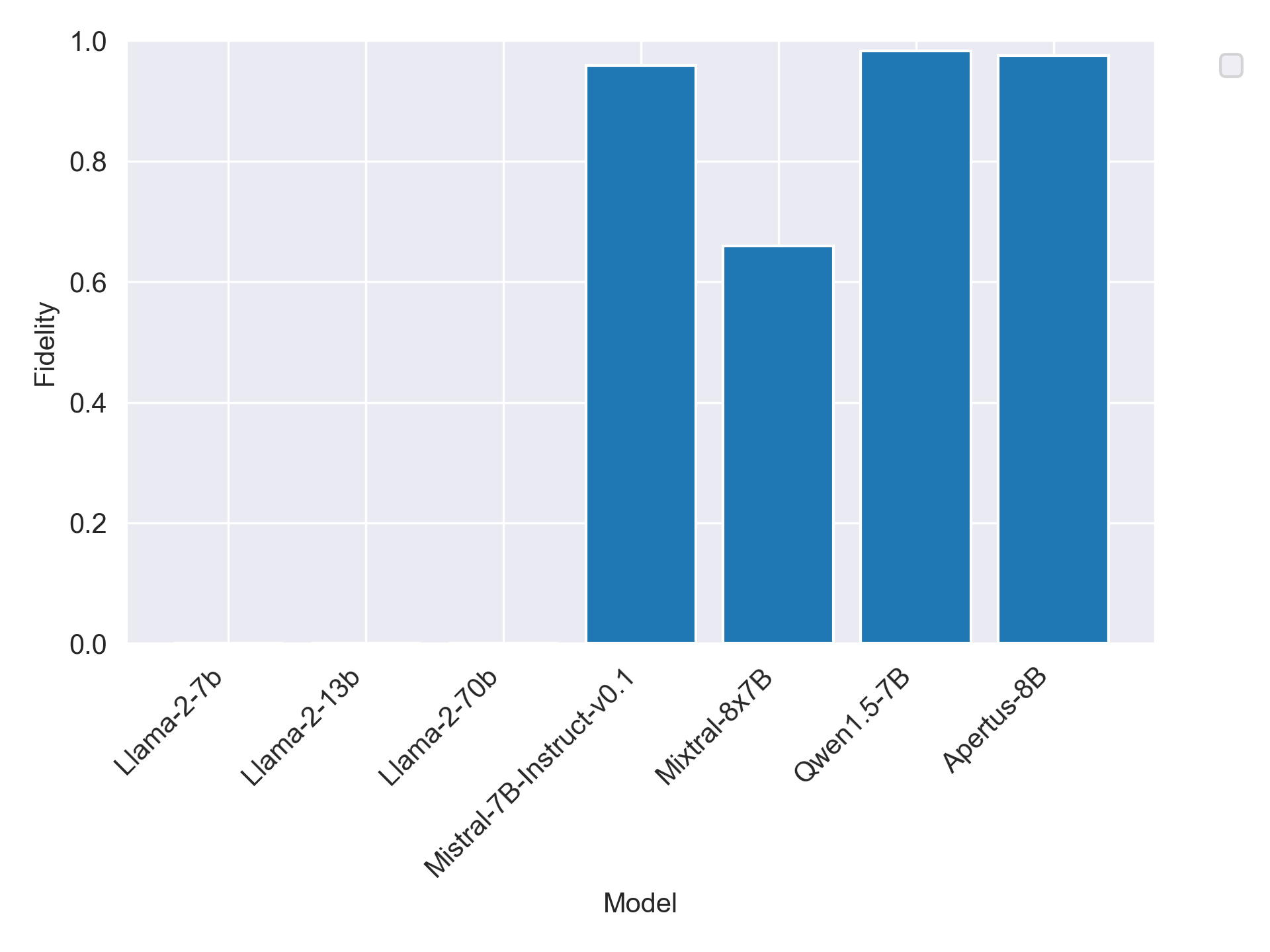}
    \caption{Chinese}
    \label{fig:fidelity_genus_Chinese}
\end{subfigure}
\vspace{1em}
\begin{subfigure}[b]{0.3\textwidth}
\centering
\includegraphics[width=\textwidth]{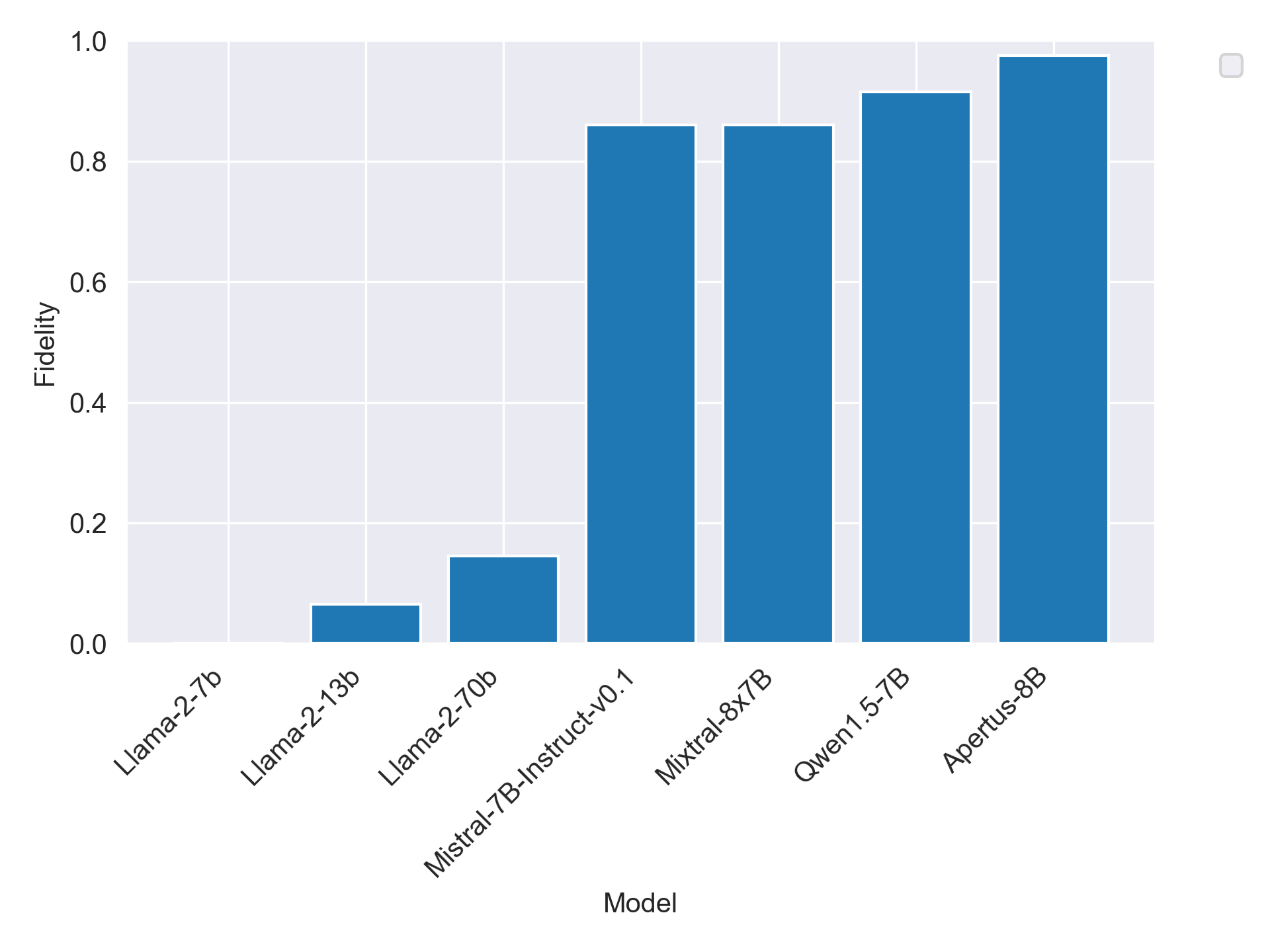}
\caption{Albanian}
\label{fig:fidelity_genus_Albanian}
\end{subfigure}
\hfill
\begin{subfigure}[b]{0.3\textwidth}
\centering
\includegraphics[width=\textwidth]{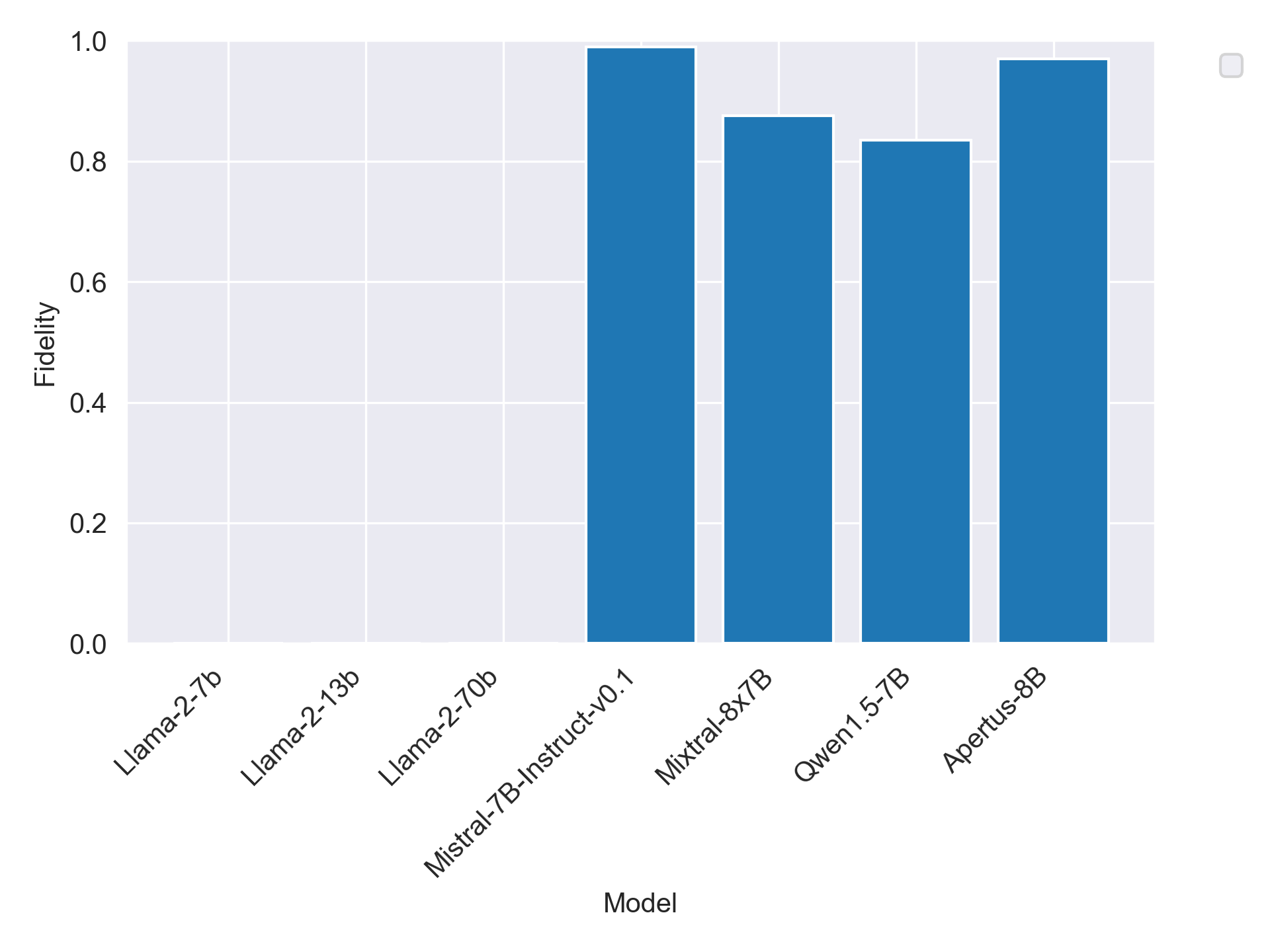}
\caption{Armenian}
\label{fig:fidelity_genus_Armenian}
\end{subfigure}
\hfill
\begin{subfigure}[b]{0.3\textwidth}
\centering
\includegraphics[width=\textwidth]{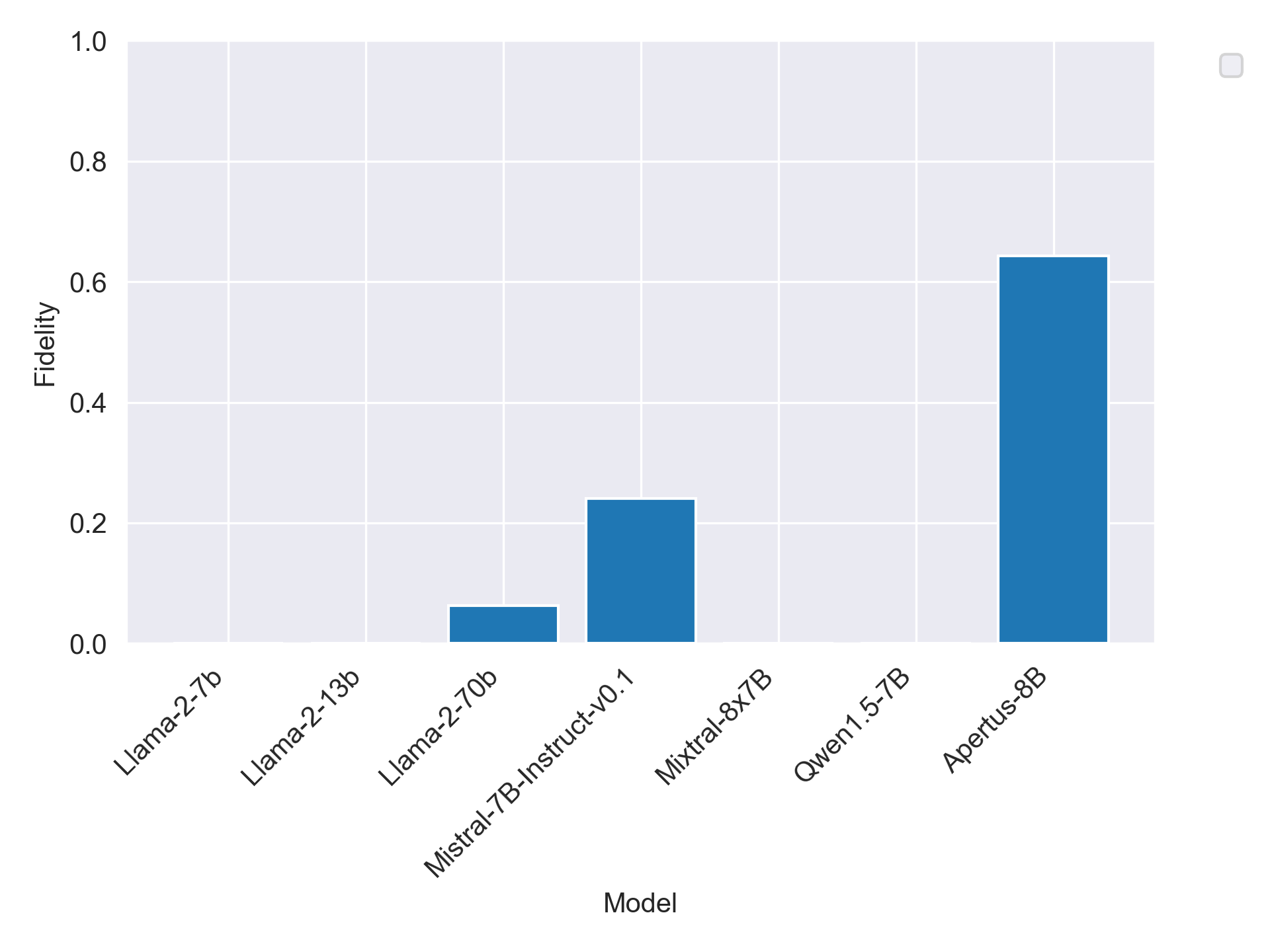}
\caption{Javanese}
\label{fig:fidelity_genus_Javanese}
\end{subfigure}
\vspace{1em}
\begin{subfigure}[b]{0.3\textwidth}
\centering
\includegraphics[width=\textwidth]{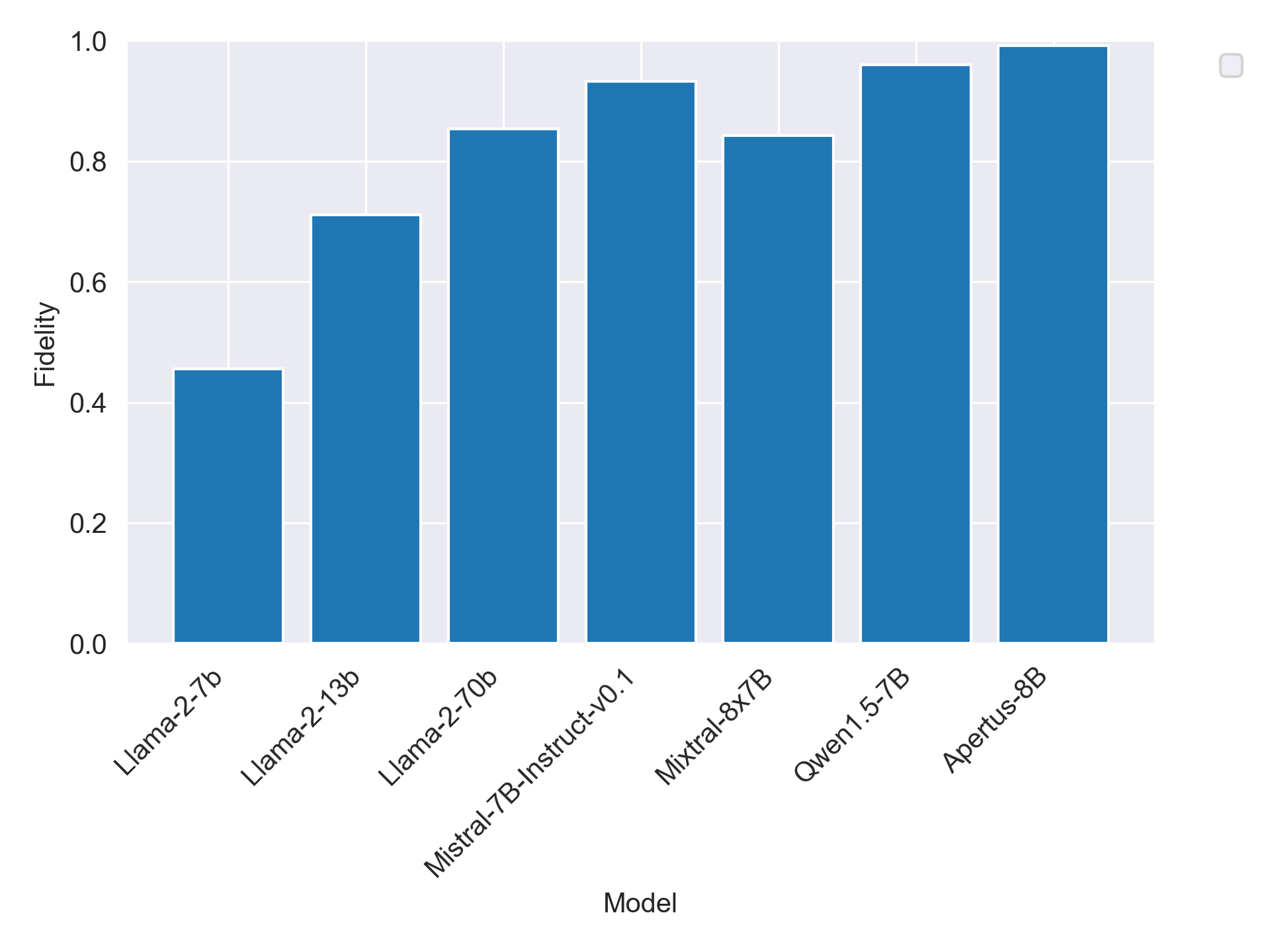}
\caption{Slavic}
\label{fig:fidelity_genus_Slavic}
\end{subfigure}
\begin{subfigure}[b]{0.3\textwidth}
\centering
\includegraphics[width=\textwidth]{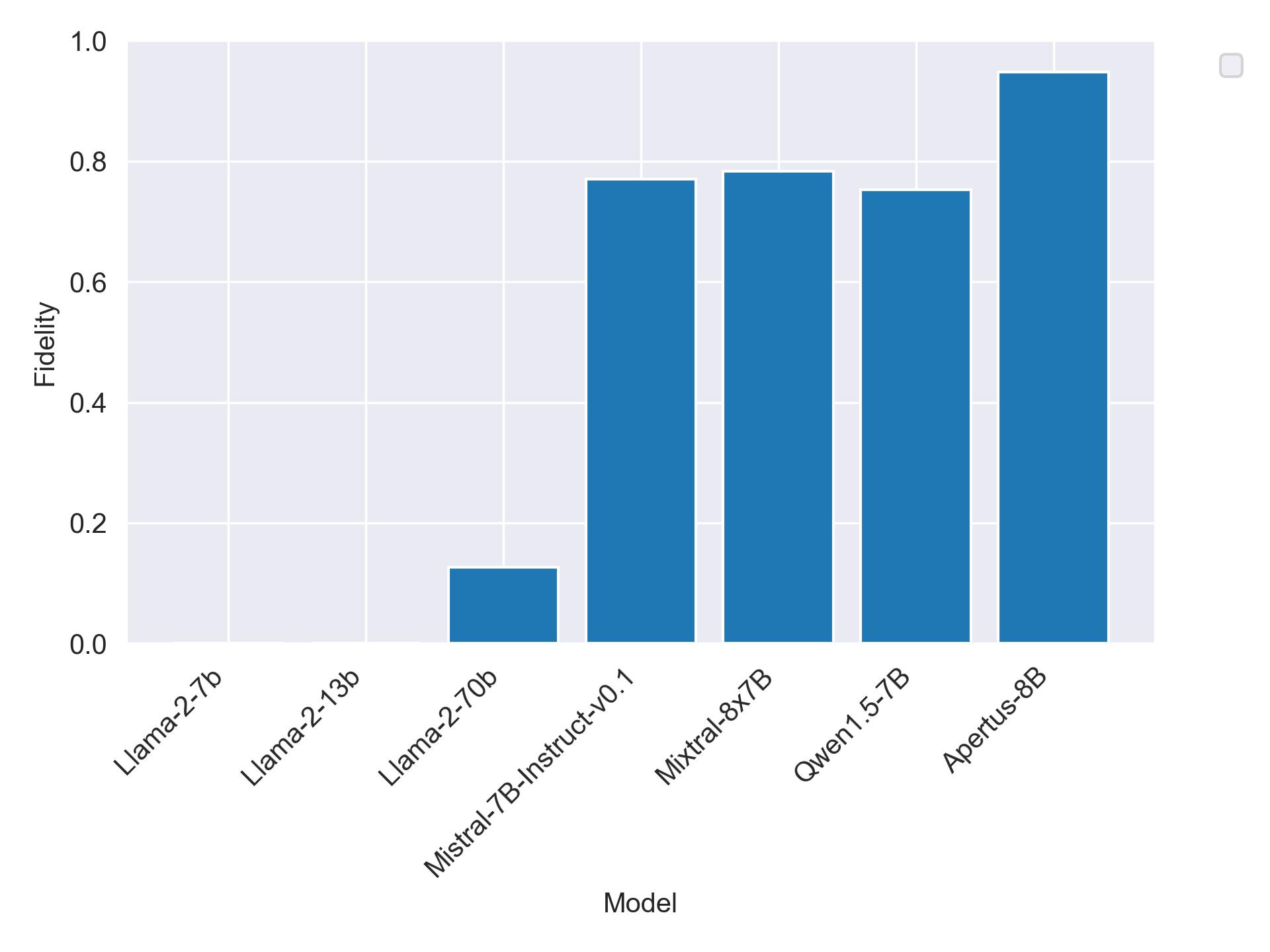}
\caption{Turkic}
\label{fig:fidelity_genus_Turkic}
\end{subfigure}
\caption{Genus-level fidelity across models.
For each representative genus, we report the proportion of model outputs that remain within the same genus as the prompt language.}
\label{fig:fidelity_genus_all}
\end{figure*}

\paragraph{Genus-level Perspective} We further provide a detailed breakdown of results for eight representative genera spanning diverse resource levels and linguistic families: Slavic, Germanic, Romance, Javanese, Albanian, Turkic, Armenian, and Chinese. We consider Javanese, Albanian and Armenian to be low-resource genera since their corresponding languages (one per genus) are low-resource languages according to the literature~\citep{nuci-etal-2024-roberta, goyal-etal-2022-flores}.

\begin{figure*}[]
    \centering
    \begin{subfigure}[b]{0.48\textwidth}
        \centering
        \includegraphics[width=\textwidth]{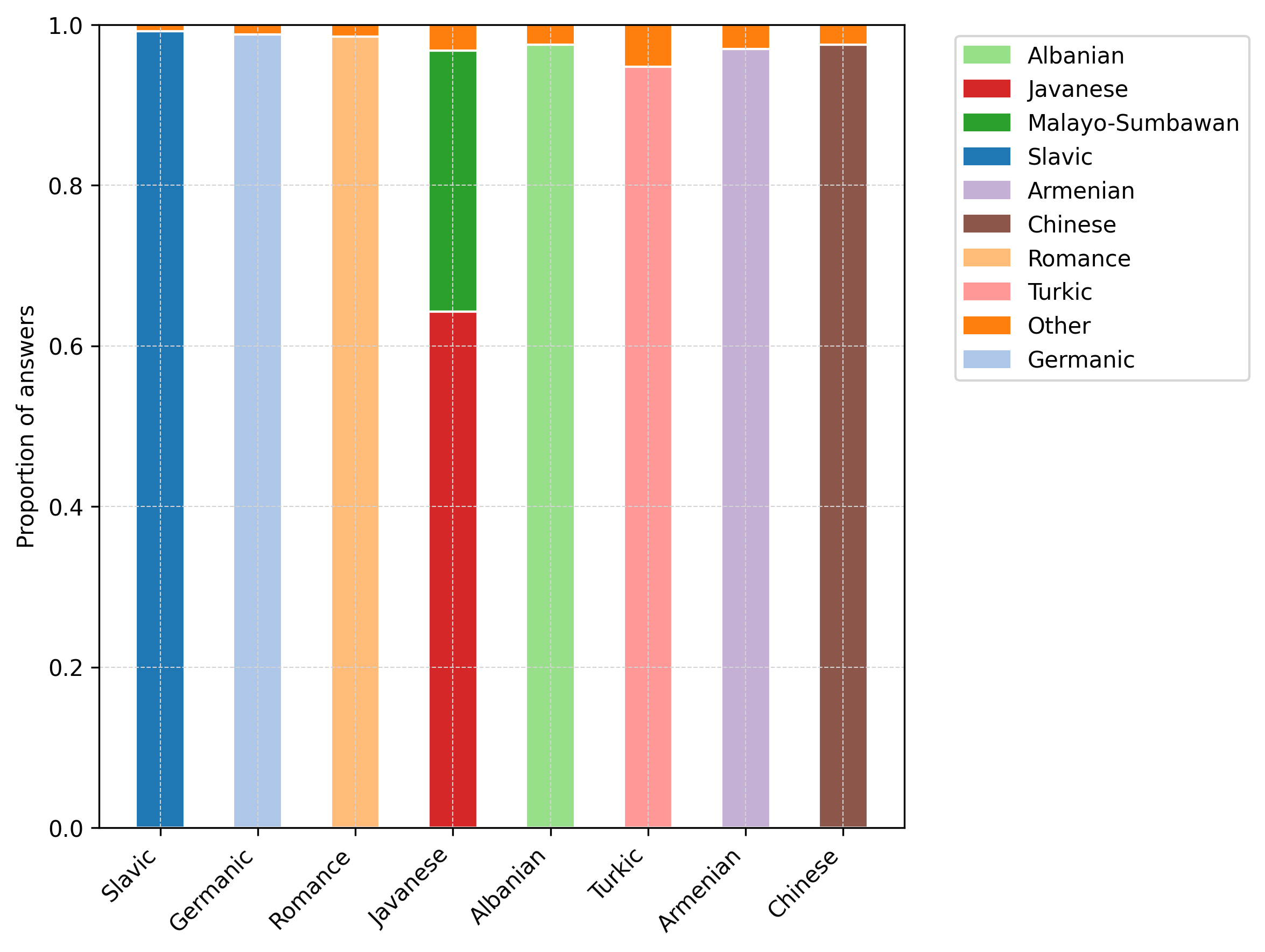}
        \caption{Apertus}
        \label{fig:fidelity_distribution_apertus}
    \end{subfigure}
    \hspace{0.02\textwidth}
    \begin{subfigure}[b]{0.48\textwidth}
        \centering
        \includegraphics[width=\textwidth]{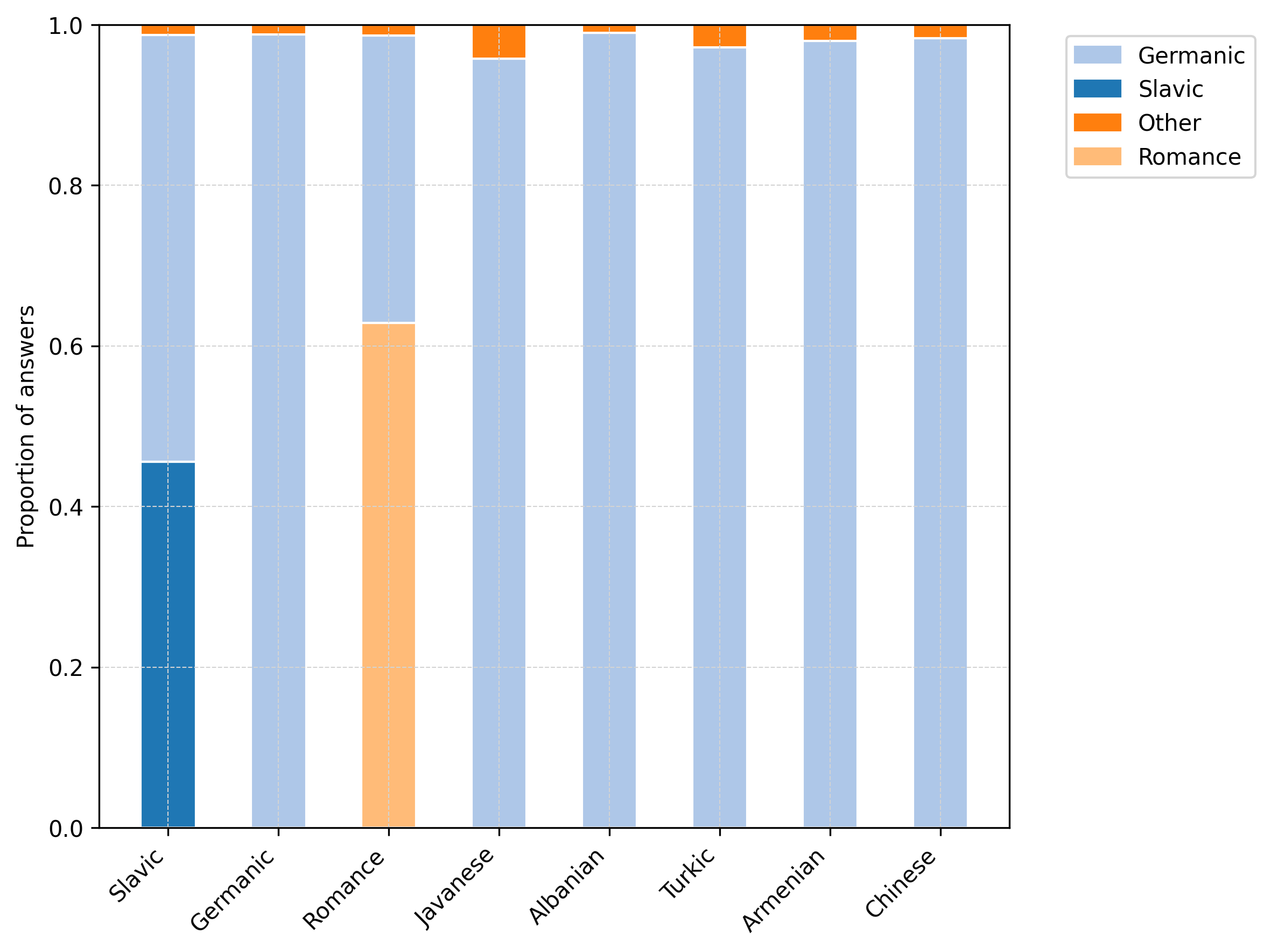}
        \caption{Llama-7b}
        \label{fig:fidelity_distribution_Llama7}
    \end{subfigure}

    \caption{Genus-level output distribution by model.
    For each prompt genus, we indicate the genus of the model’s generated response. 
    Remaining models are reported in Appendix~\ref{app:detail_output_model}.}
    \label{fig:fidelity_distribution_main}
\end{figure*}

Results for these eight genera are shown in the Figure~\ref{fig:fidelity_genus_all}. 
We additionally show Apertus-8B and Llama-2-7b (having the highest contrasted performances) in Figures~\ref{fig:fidelity_distribution_apertus} and~\ref{fig:fidelity_distribution_Llama7}, with the other models detailed in Appendix~\ref{app:detail_output_model}. 

Across all models, Germanic languages consistently achieve high genus fidelity ($\geq 0.75$). A similar trend is observed for Romance languages, except for the smallest Llama model. Even lower-resource languages within these genera benefit from this stability (except from Llama models which show systematic bias towards English and thus Germanic genus as a fallback). 

For Chinese languages, fidelity significantly drops for Llama models and Mixtral. In particular, Llama models again show a pronounced Germanic bias in over 60\% of non-faithful cases. This pattern suggests weaker multilingual competence for these genera, with models reverting to training-dominant languages.

Fidelity to Slavic languages overall remains strong (over 0.8) apart from two smaller models from Llama family.

For low-resource genera, fidelity varies substantially across models. While models maintain a minimal fidelity to Albanian (even Llama), only non-Llama models preserve fidelity for Armenian, Mistral and Apertus achieving particularly high fidelity scores. In contrast, for Javanese most models struggle and even Apertus barely reaches 0.6 fidelity score.

Turkic languages also exhibit complex patterns. Mistral and Qwen maintain a fielity above $0.75$ , whereas Llama produces Germanic outputs in over 80\% of cases. Detailed inspection reveals substantial intra-genus variation: Turkish prompts yield relatively faithful responses, while Kazakh frequently triggers English outputs.
This disparity likely reflects multiple factors: resource imbalance (Turkish being better represented in training data), script effects, and contact phenomena. 

\section{Genus switch}\label{s:switch}


If an LLM answers a question correctly in one language, is it more likely to answer correctly when the same question is posed in another language of the same genus? We investigate whether genus consistency facilitates knowledge transfer across languages.

\subsection{Methology}
If a model demonstrates knowledge by answering correctly in one language, changing only the prompt language should not impede correct responses --assuming sufficient multilingual competence. We test whether genealogical proximity preserves this knowledge consistency better than genealogically distant language pairs.

\paragraph{Setup}
Using MultiQ, we identify questions answered correctly in a source language, then evaluate the same questions across all available target languages. This controlled design isolates language effects from knowledge availability, since the model has already demonstrated requisite knowledge.

\paragraph{Metrics}
We compute SwitchScores measuring the proportion of questions answered correctly in target genus $g_t$ given correct answers in source genus $g_i$:

\begin{equation}
\text{SwitchScore}(g_i, g_t) =
\frac{\mid\mathcal{Q}g_i,g_t\mid}{\mid\mathcal{Q}g_{i}\mid}
\end{equation}
\noindent where $\mathcal{Q}_{g_i,g_t}$
 represents questions answered correctly in both genera, and $\mathcal{Q}_{g_i}$
 represents questions answered correctly in the source genus.

We distinguish:

\begin{itemize}
    \item SwitchScore-In (within the same genus): \text{SwitchScore}($g_i$, $g_i$) - within-genus consistency

    \item SwitchScore-Out (outside of input genus): Average performance when switching to other genera (different from $g_i$)
\end{itemize}

\paragraph{Question selection}\label{ss:quest_selct}

The original MultiQ evaluation used the complete dataset across all languages. However, to ensure that observed differences genuinely reflect language effects rather than artifacts of question difficulty or translation quality, we construct a filtered subset optimized for cross-genus comparison.

More specifically, a question is retained if it is answerable across the compared genera, i.e., the model produces a correct answer in at least one language within each genus. This prevents biases arising from inherently unanswerable questions.

Moreover, we only keep languages where the model achieves at least a minimal number of correct answers $N_c$ to ensure statistical reliability. We use $N_c$ values of 20, 50 and 100.

\paragraph{Resulting Dataset Characteristics}
Our filtering process yields a curated dataset optimized for genealogical analysis while maintaining the linguistic diversity required for robust conclusions.
Table~\ref{tab:size_gen_q_threshold} presents the resulting dataset size under different filtering thresholds.
With this approach, we aim at prioritizing interpretability and statistical validity over dataset size, ensuring that our genealogical findings reflect genuine linguistic patterns.

\begin{figure*}[!t]
    \centering
    \begin{subfigure}[b]{0.48\textwidth}
        \centering
        \includegraphics[width=\textwidth]{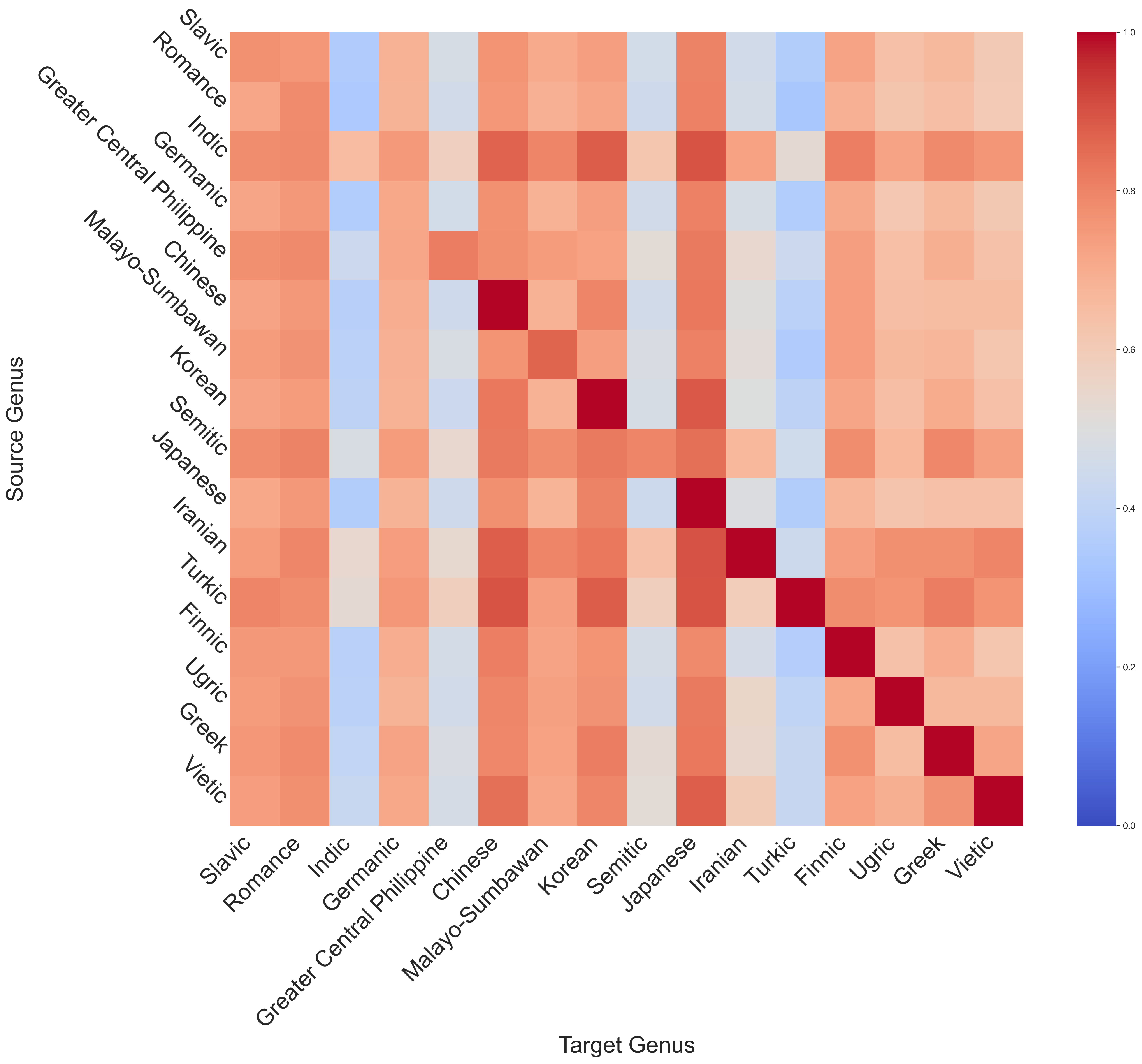}
        \caption{Llama-70b, threshold = 50}
        \label{fig:switchscore_Llama70_50_main}
    \end{subfigure}
    \hspace{0.02\textwidth}
    \begin{subfigure}[b]{0.48\textwidth}
        \centering
        \includegraphics[width=\textwidth]{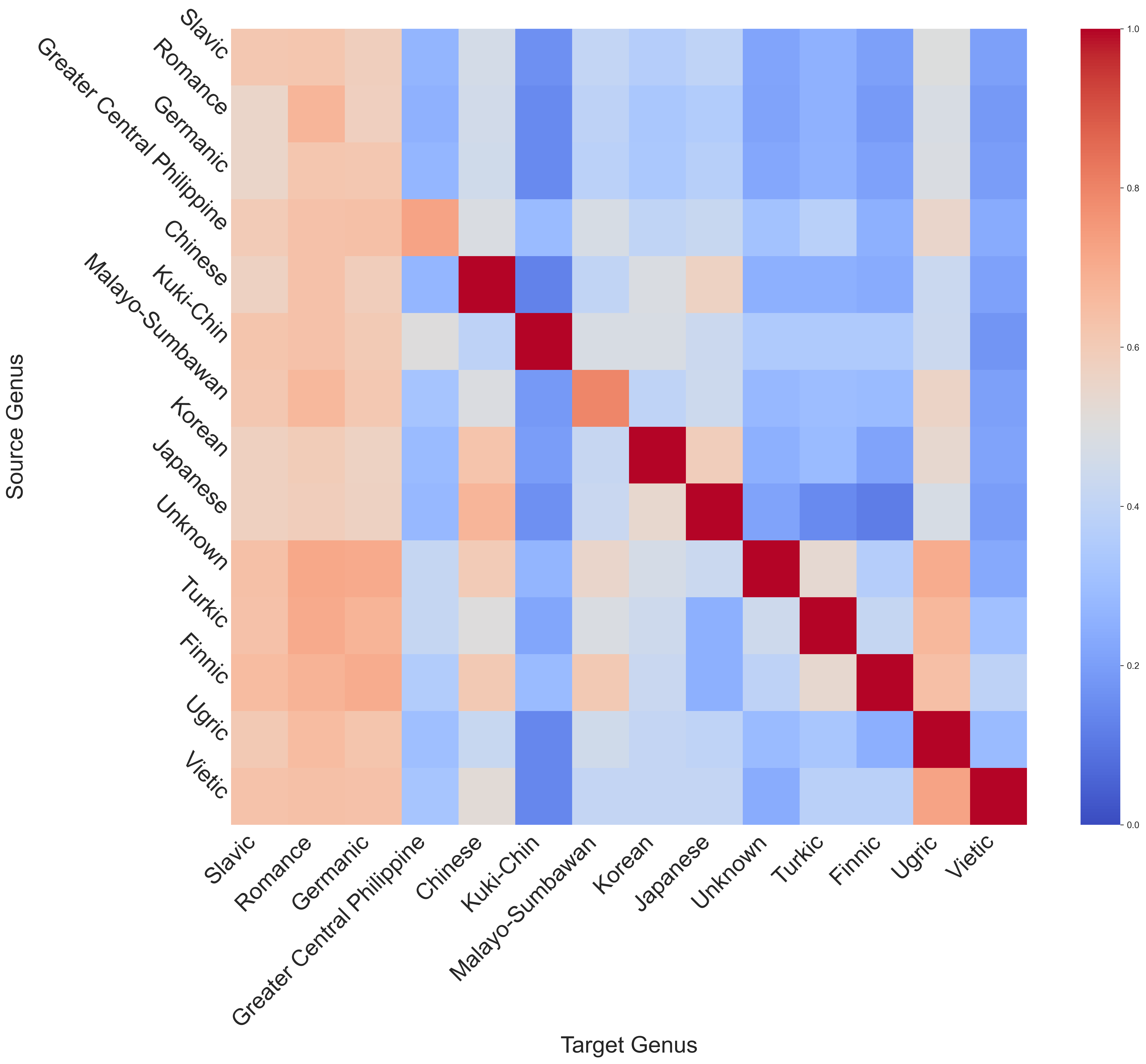}
        \caption{Mistral-7B, threshold = 20}
        \label{fig:switchscore_Mistral_20_main}
    \end{subfigure}

    \vspace{0.5em} 

    \begin{subfigure}[b]{0.48\textwidth}
        \centering
        \includegraphics[width=\textwidth]{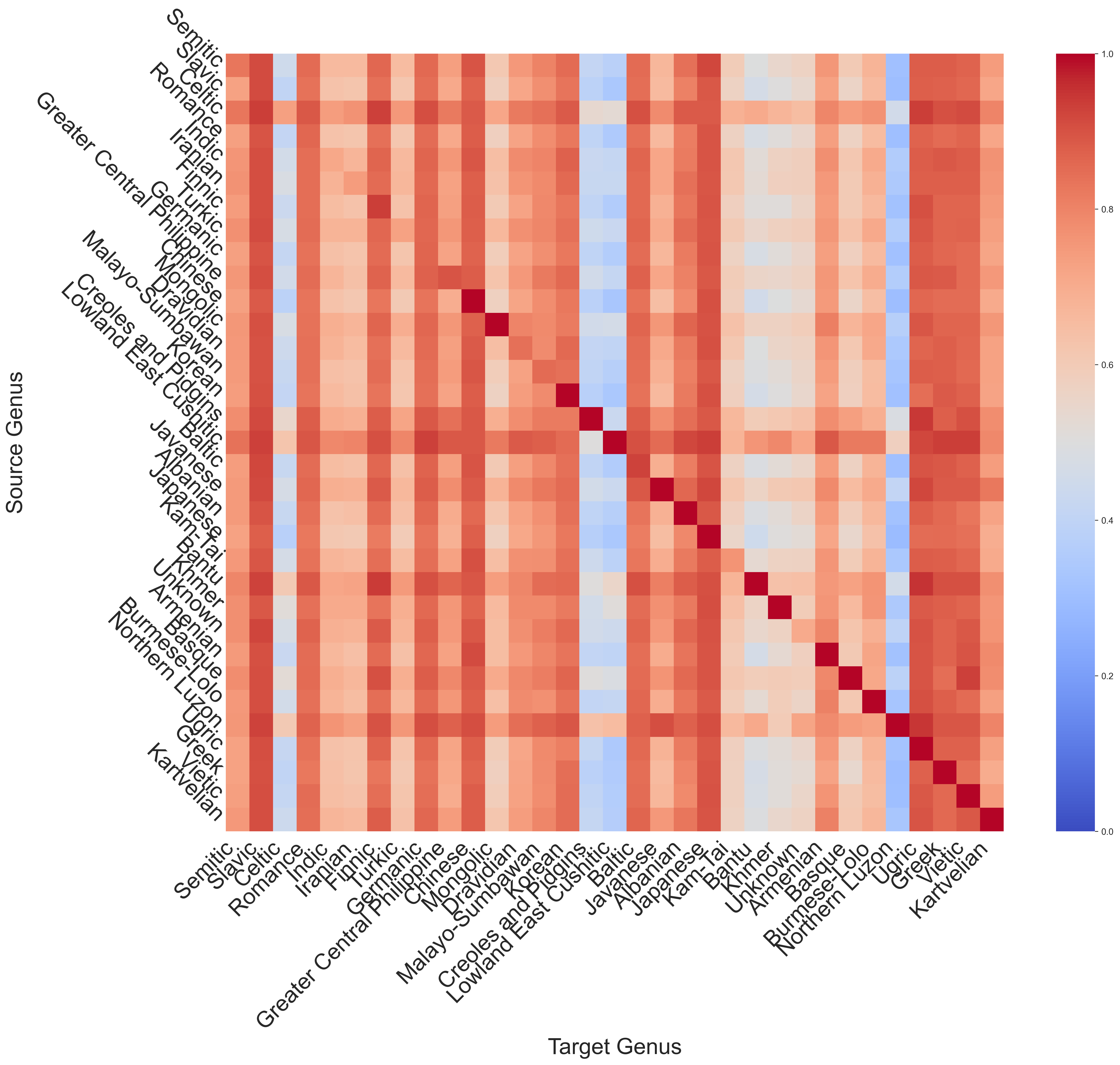}
        \caption{Apertus-8B, threshold = 50}
        \label{fig:switchscore_apertus_50_main}
    \end{subfigure}
    \hspace{0.02\textwidth}
    \begin{subfigure}[b]{0.48\textwidth}
        \centering
        \includegraphics[width=\textwidth]{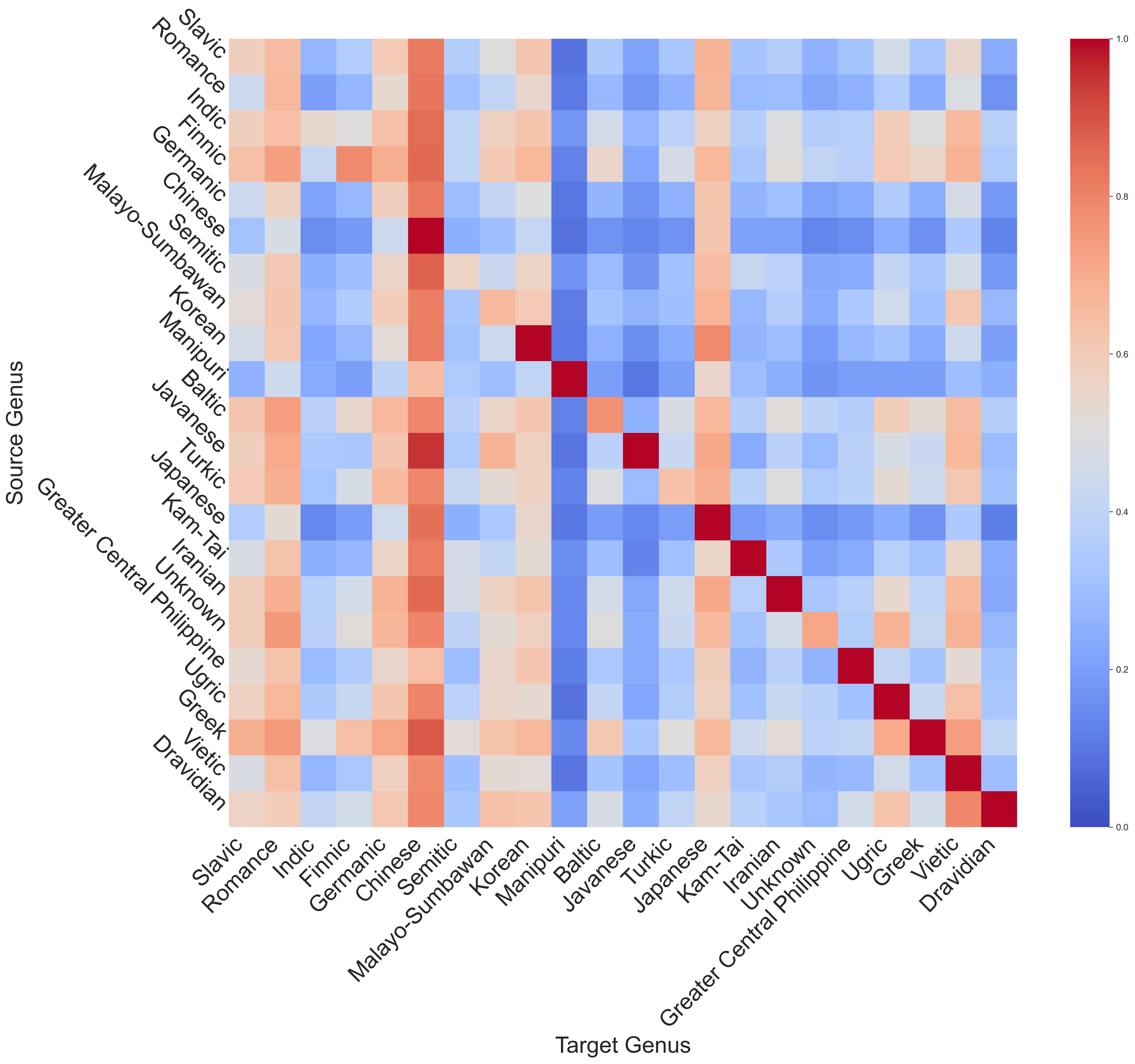}
        \caption{Qwen-1.5-7B, threshold = 20}
        \label{fig:switchscore_Qwen_20_main}
    \end{subfigure}

    \caption{Switchscores across models and thresholds. 
    Each subfigure shows the switchscore distribution for one model at the specified threshold. As can be seen from red/blue-column patterns, the performance critically depends on the target genus.}
    \label{fig:switchscores_all}
\end{figure*}

\begin{table*}[!t]
\centering
\begin{tabular}{l
                cc cc cc cc cc cc cc}
\toprule
$N_c$ & 
\multicolumn{2}{c}{\textbf{Llama-7B}} & 
\multicolumn{2}{c}{\textbf{Llama-13B}} & 
\multicolumn{2}{c}{\textbf{Llama-70B}} & 
\multicolumn{2}{c}{\textbf{Mistral-7B}} & 
\multicolumn{2}{c}{\textbf{Mixtral-8x7}} & 
\multicolumn{2}{c}{\textbf{Qwen-7B}} & 
\multicolumn{2}{c}{\textbf{Apertus-8B}} \\
\cmidrule(lr){2-3} \cmidrule(lr){4-5} \cmidrule(lr){6-7} 
\cmidrule(lr){8-9} \cmidrule(lr){10-11} \cmidrule(lr){12-13} \cmidrule(lr){14-15}
\\
 & \# q & \# g & \# q & \# g & \# q & \# g & \# q & \# g & \# q & \# g & \# q & \# g & \# q & \# g \\
\midrule 
20  & 7 & 26 &5 & 23 &4 & 24 &8 & 33 &3 & 14 &3 & 22 & \textbf{14} & \textbf{44} \\
50  & 6 & 16 &4 & 17 &3 & 15 &7 & 20 &3 & 8 &2 & 9 & \textbf{14} & \textbf{33}   \\
100 &5 & 11 &3 & 7 &1 & 6 &5 & 11 &2 & 3 &1 & 4 & \textbf{13} & \textbf{27} \\
\bottomrule
\end{tabular}
\caption{Number of questions (\# q, expressed in thousands) and number of genera (\# g) remaining after applying different filtering thresholds for each model.}
\label{tab:size_gen_q_threshold}
\end{table*}

\subsection{Results}

Global SwitchScores appear in Table~\ref{tab:genus_switch_score}. Detailed genus-level scores are shown in Figures~\ref{fig:switchscores_all} for Llama-2-70b (\ref{fig:switchscore_Llama70_50_main}), Mistral-7B(\ref{fig:switchscore_Mistral_20_main}), Apertus-8B (\ref{fig:switchscore_apertus_50_main}) and Qwen-1.5-7B(\ref{fig:switchscore_Qwen_20_main}), with additional results in Appendix~\ref{app:detail_switchscores}.

\begin{table}[!h]
\centering
\begin{tabular}{lll}
\toprule
Model & Switch-In & Switch-Out \\
\midrule
Llama-2-7b  & 88.9 & 49.4 \\
Llama-2-14b & 82.8 & 51.6  \\
Llama-2-70b & 86.3 & 54.0  \\
Mixtral-8x7B  & 83.6 & 47.7  \\
Mistral-7B & 88.8  & 41.8 \\
Qwen1.5-7B-Chat  & 84.1  & 42.0  \\
Apertus-8B & \textbf{90.4} & \textbf{60.6} \\
\bottomrule
\end{tabular}
\caption{Switch scores by model. Obtained with a threshold of 20.}
\label{tab:genus_switch_score}
\end{table}

All models show substantially higher knowledge consistency within genera (80-90\%) compared to cross-genus transfers (40-50\%). This 35-40 percentage point advantage demonstrates that genealogical relatedness significantly facilitates knowledge preservation.

\paragraph{Detailed Switchscores}

A key finding is that performance depends critically on the target genus rather than the source (notice the red/blue column pattern across Figures~\ref{fig:switchscores_all}). Well-resourced genera (Germanic, Romance) serve as robust targets regardless of source, while poorly resourced genera (e.g., Kuki-Chin) yield degraded performance even from high-resource sources.

Moreover, results are asymmetric: for Llama-70b switching from Javanese (Austronesian) to Germanic maintains high accuracy, whereas the reverse direction shows substantial degradation. This suggests that target language representation in training data dominates genealogical effects when resources are scarce.

\paragraph{Genealogical Boundaries} 
Despite overall genus-level patterns, genealogical classification does not perfectly predict transfer success. Within Indo-European, Indic and Slavic genera exhibit markedly different behaviors despite shared family membership. Similarly, script overlap (Latin, Cyrillic) provides no guarantee of stable transfer performance.
These exceptions highlight that while genealogical relatedness provides a useful organizational principle for understanding multilingual LLM behavior, it competes with training data distribution, script similarity, and other linguistic factors in determining cross-lingual knowledge consistency.

\section{Conclusions}\label{s:ccl}

In this paper, we examined the genealogical sensitivity of Large Language Models through a genus-level analysis, extending the work of \citet{holtermann-etal-2024-evaluating}. We found that LLMs exhibit higher fidelity and knowledge consistency within genealogical boundaries, but this effect is largely mediated by training resource availability. Distinct multilingual strategies also emerged across model families, with models defaulting to Germanic languages and others adopting more nuanced behaviors. Overall, our findings indicate that resource distribution, rather than genealogical structure, remains the primary driver of multilingual performance.


\section*{Impact statement} 

As we are witnessing the progressive usage of LLMs, also for the scopes of generating different benchmarks, we would like to remind that even these less-resource intensive activities contribute to high energy consumption and carbon emissions. To give our small contribution to the AI sustainability, we opted to use existing benchmark and intervene only as needed. We hope that this can inspire other LLM-related research to leverage existing resources at least equally optimally. 

\section{Bibliographical References}\label{sec:reference}

\bibliographystyle{lrec2026-natbib}
\bibliography{lrec2026-example}

\label{lr:ref}
\bibliographystylelanguageresource{lrec2026-natbib}

\appendix

\clearpage
\section{Genus output detail}\label{app:detail_genus_output}

\subsection{Detail per model}\label{app:detail_output_model}

\begin{figure*}[t]
\centering

\begin{subfigure}[b]{0.45\textwidth}
\centering
\includegraphics[width=\textwidth]{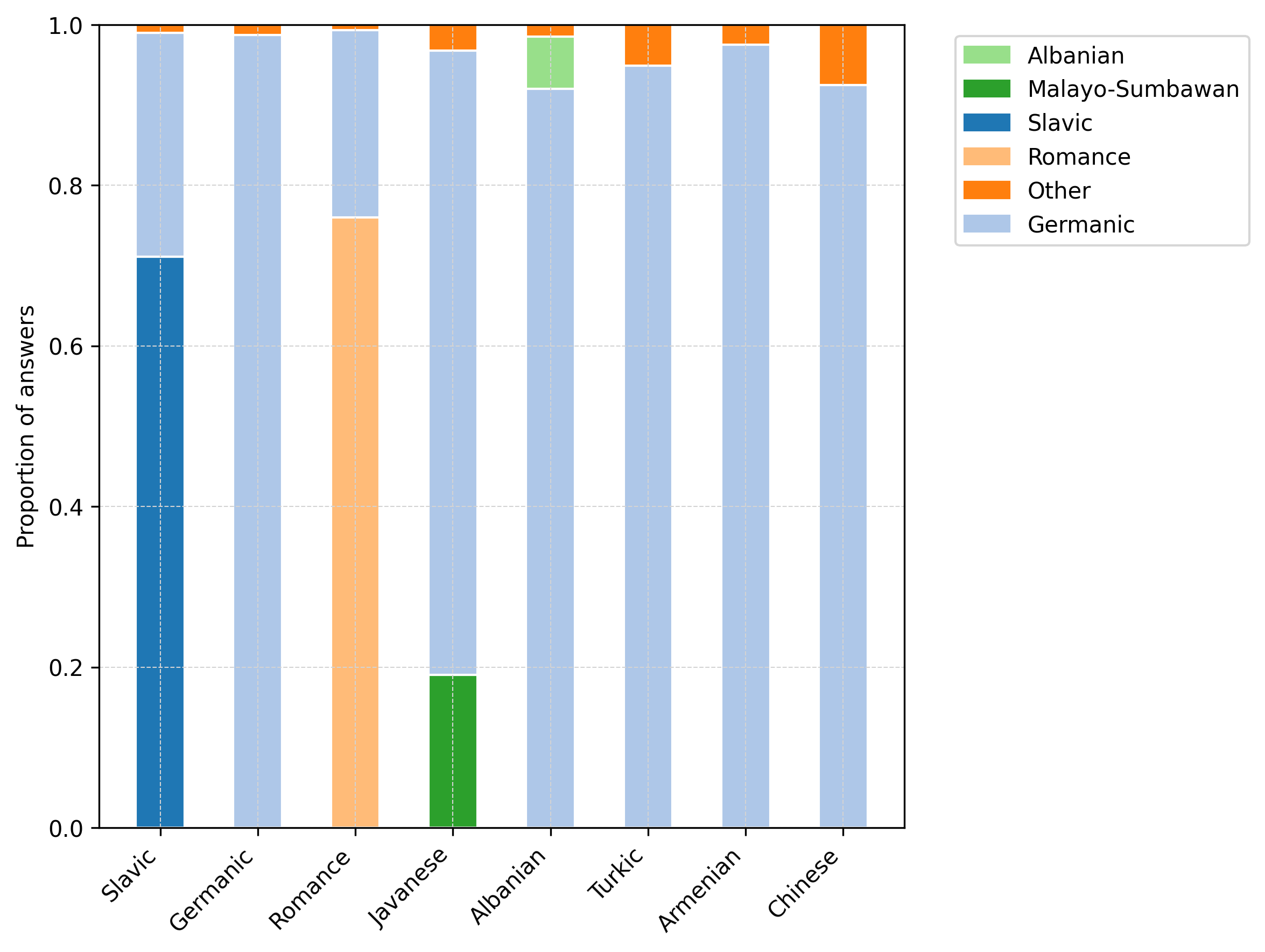}
\caption{Llama-13b}
\label{fig:fidelity_distribution_Llama-13b}
\end{subfigure}
\hfill
\begin{subfigure}[b]{0.45\textwidth}
\centering
\includegraphics[width=\textwidth]{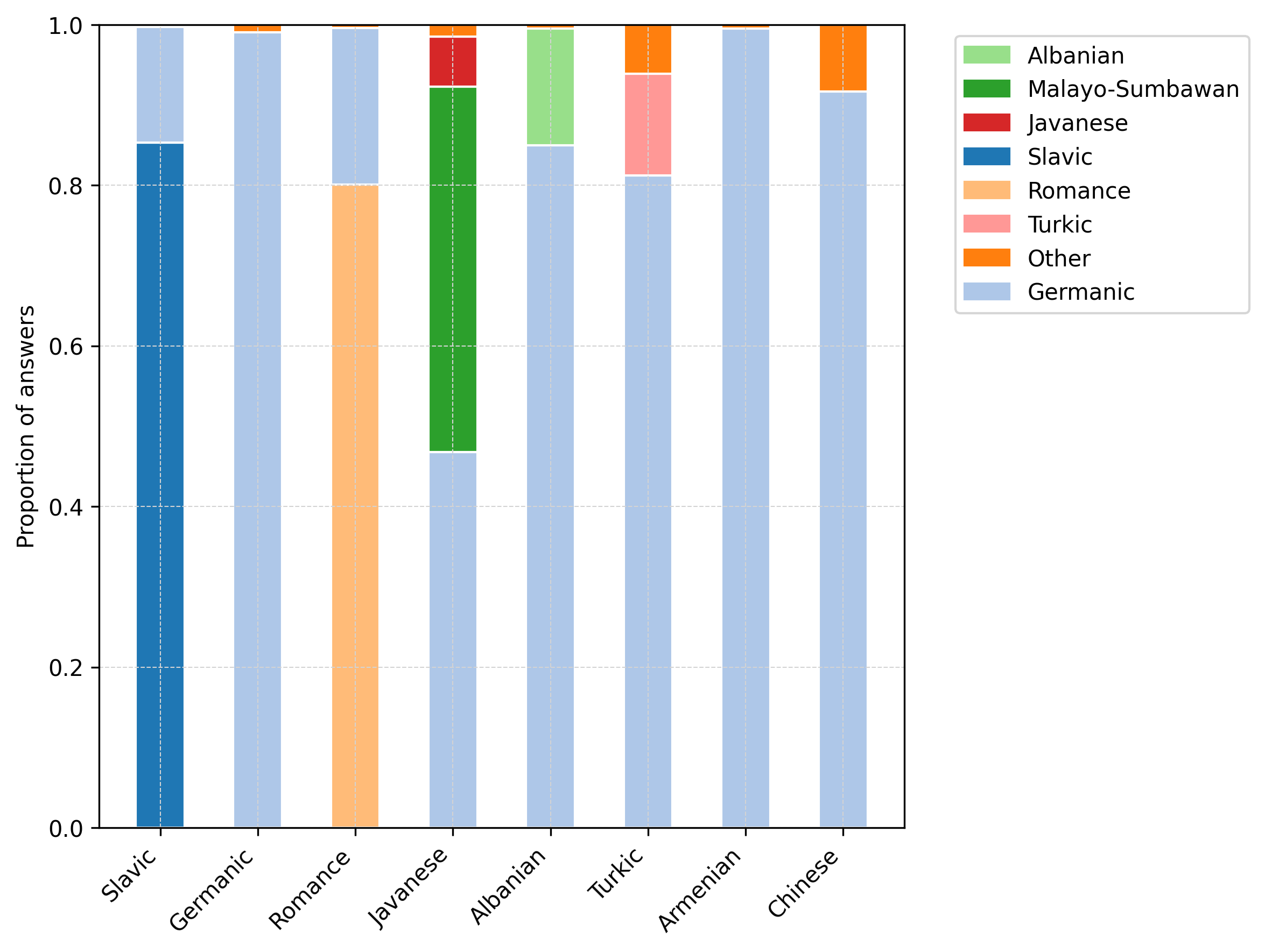}
\caption{Llama-70b}
\label{fig:fidelity_distribution_Llama-70b}
\end{subfigure}

\vspace{1em}

\begin{subfigure}[b]{0.45\textwidth}
\centering
\includegraphics[width=\textwidth]{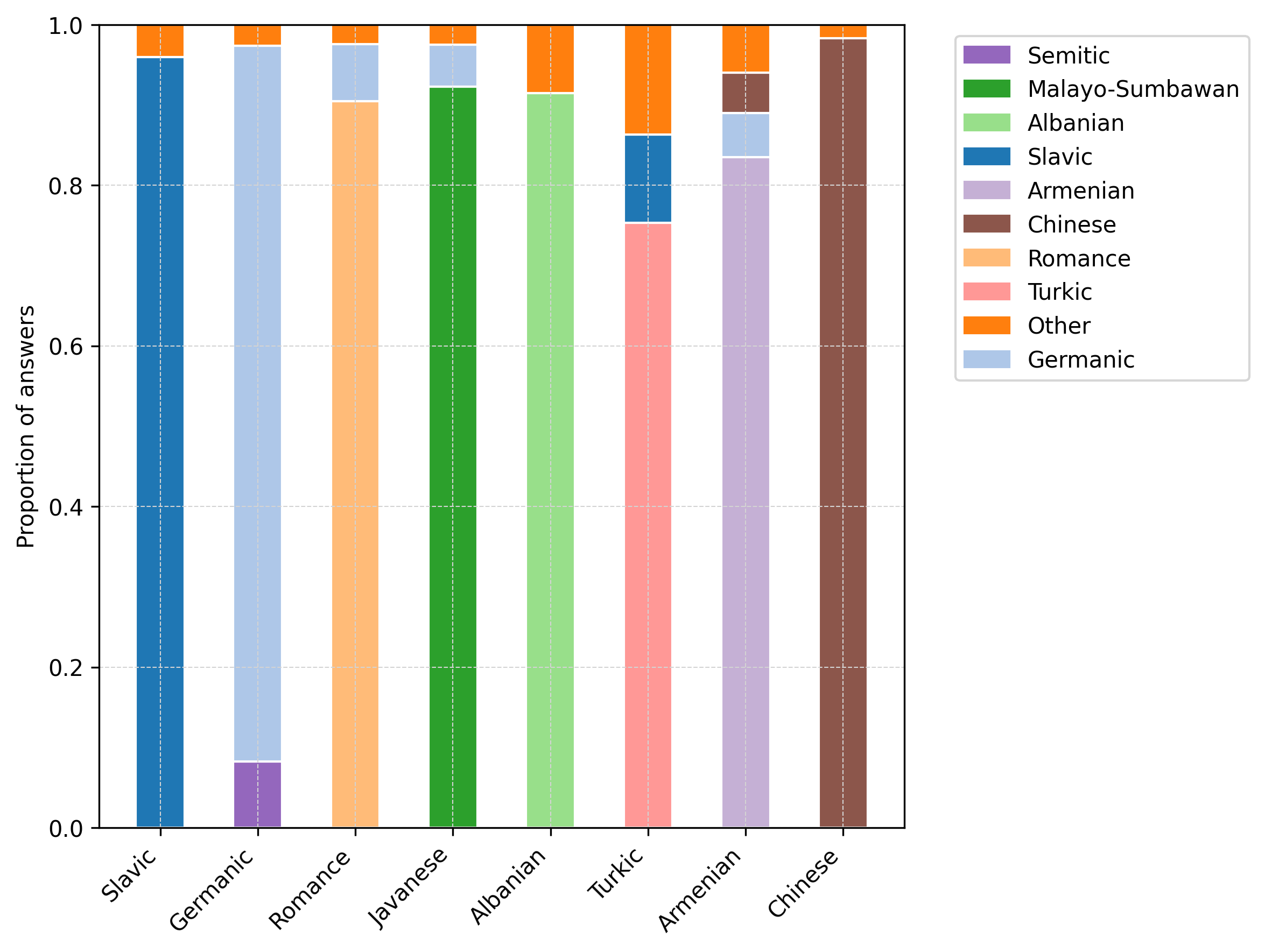}
\caption{Qwen1.5-7B}
\label{fig:fidelity_distribution_Qwen}
\end{subfigure}
\hfill
\begin{subfigure}[b]{0.45\textwidth}
\centering
\includegraphics[width=\textwidth]{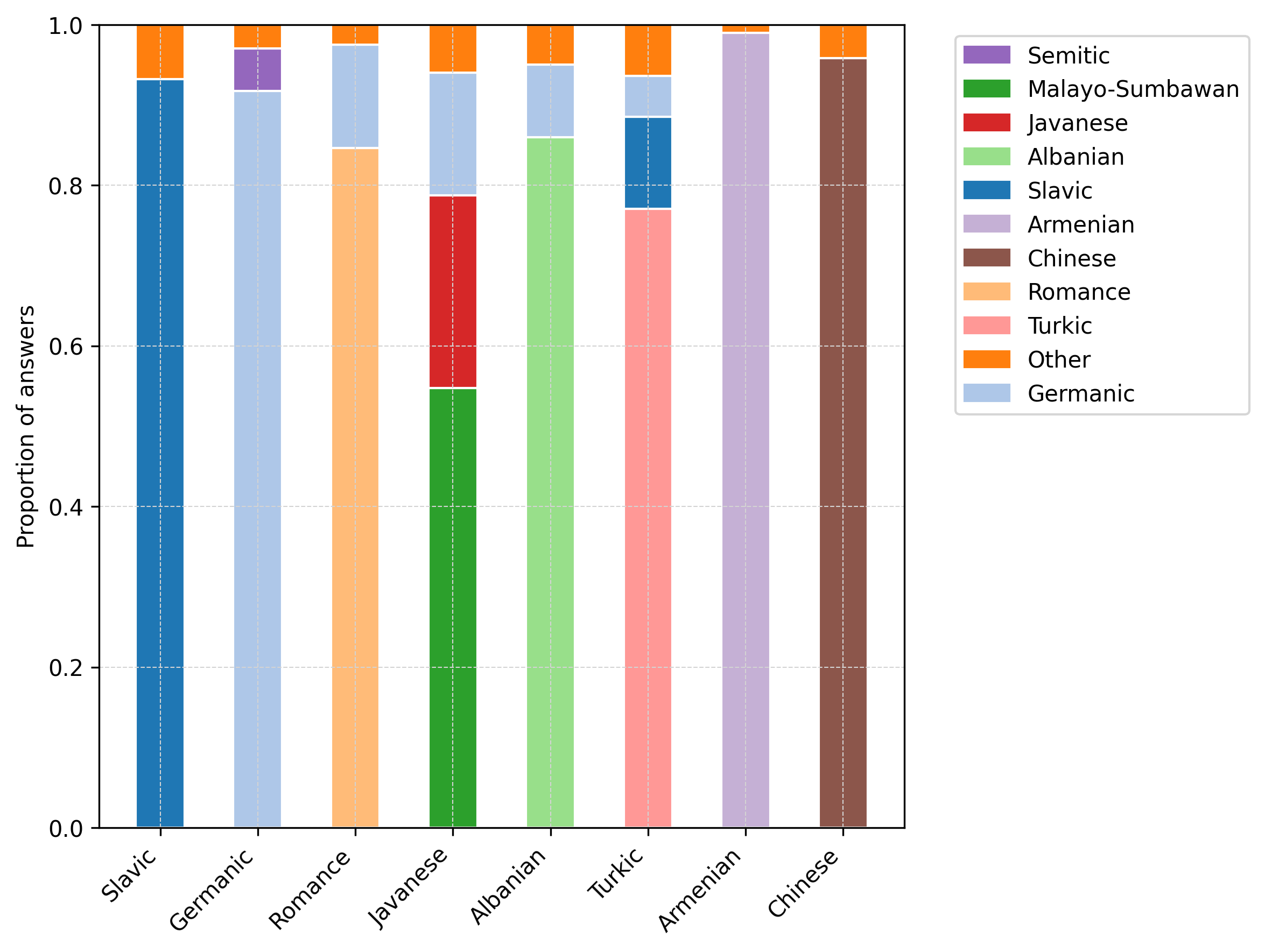}
\caption{Mistral-7B}
\label{fig:fidelity_distribution_Mistral}
\end{subfigure}

\vspace{1em}

\begin{subfigure}[b]{0.45\textwidth}
\centering
\includegraphics[width=\textwidth]{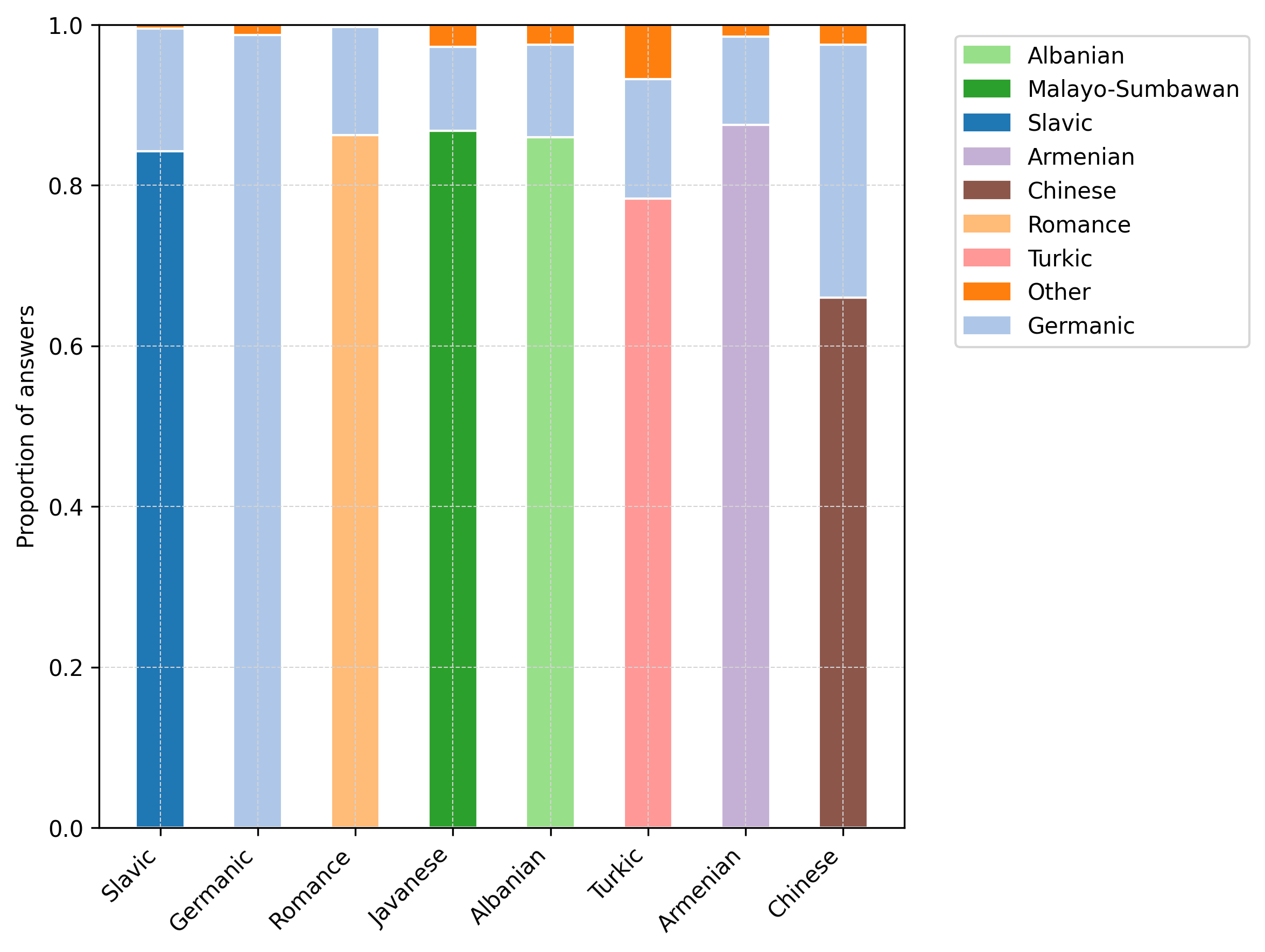}
\caption{Mixtral-8x7B}
\label{fig:fidelity_distribution_Mixtral-8x7B}
\end{subfigure}

\caption{Genus-level output distribution by model.
For each prompt genus, we indicate the genus of the model’s generated response.}
\label{fig:fidelity_distribution_app}
\end{figure*}

The details of output genera for eight selected genera per model can be seen in Figure~\ref{fig:fidelity_distribution_app}.

\clearpage

\section{Switchscores}\label{app:detail_switchscores}

\begin{figure}[!h]
    \centering
    \begin{subfigure}[b]{0.45\textwidth}
        \centering
        \includegraphics[width=0.7\textwidth]{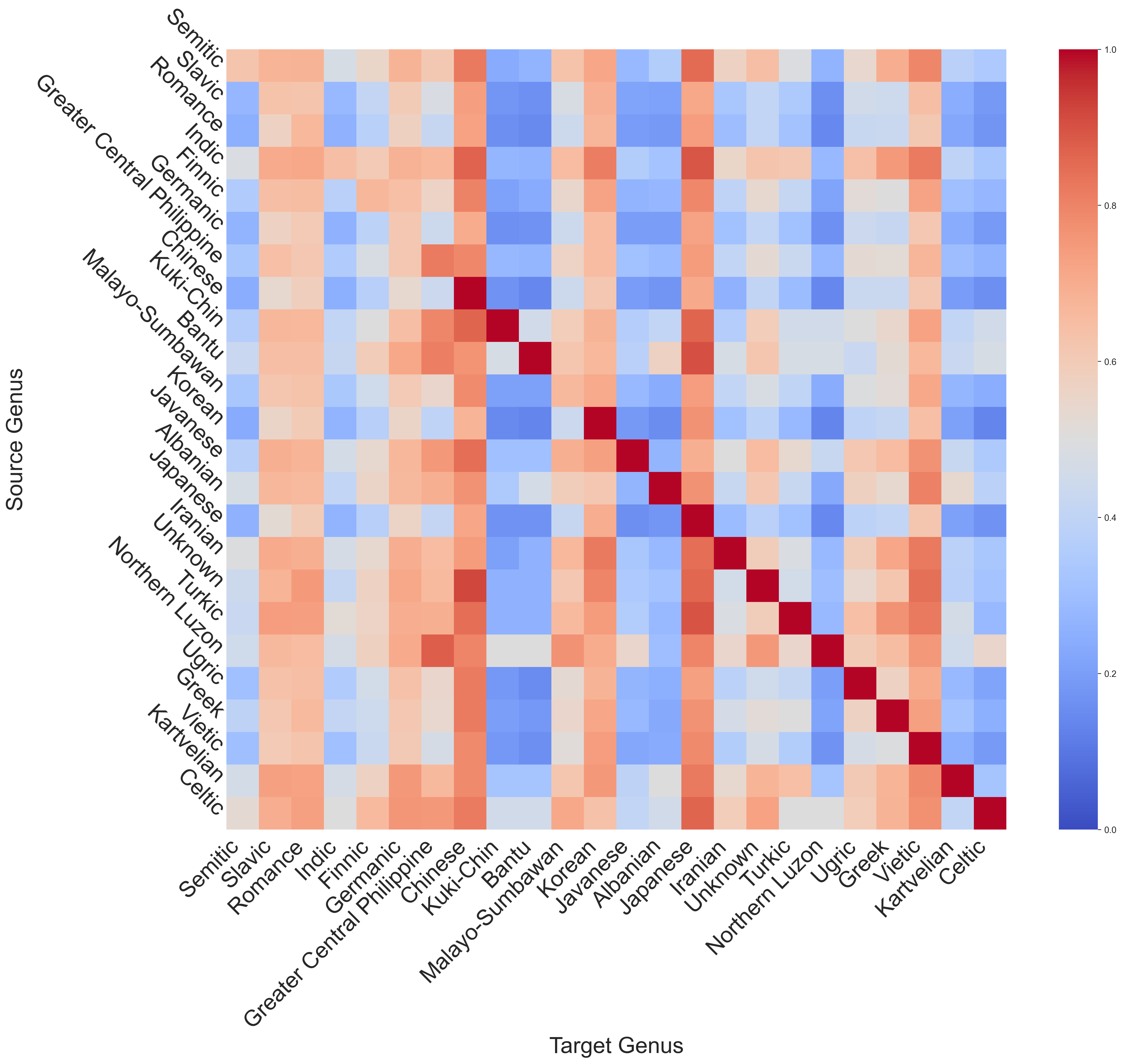}
        \caption{Switchscores of model Llama-7b, threshold 20.}
        \label{fig:switchscore_Llama7_th20}
    \end{subfigure}
    \hfill
    \begin{subfigure}[b]{0.45\textwidth}
        \centering
        \includegraphics[width=0.7\textwidth]{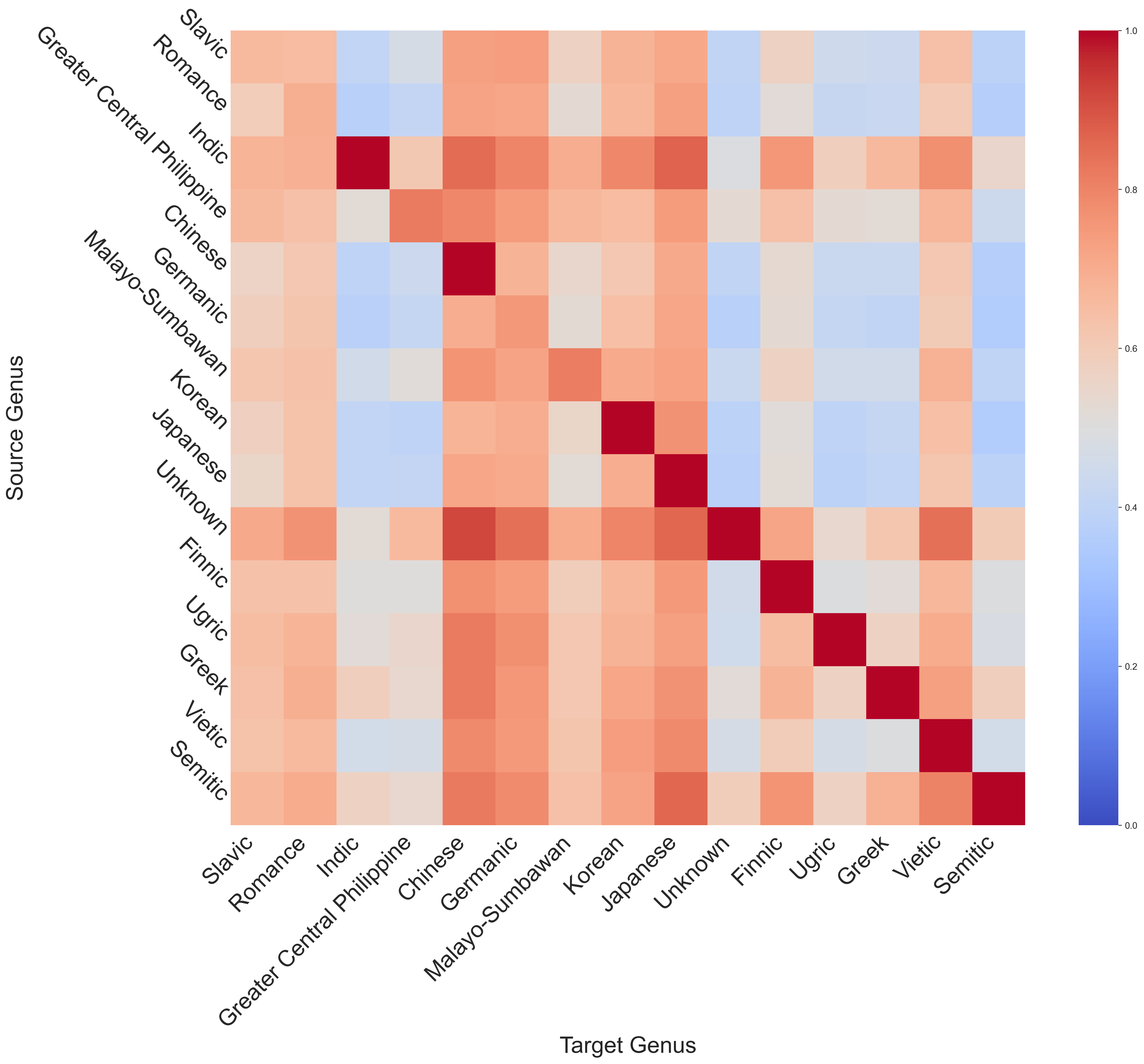}
        \caption{Switchscores of model Llama-7b, threshold 50.}
        \label{fig:switchscore_Llama7_th50}
    \end{subfigure}
    \hfill
    \begin{subfigure}[b]{0.45\textwidth}
        \centering
        \includegraphics[width=0.7\textwidth]{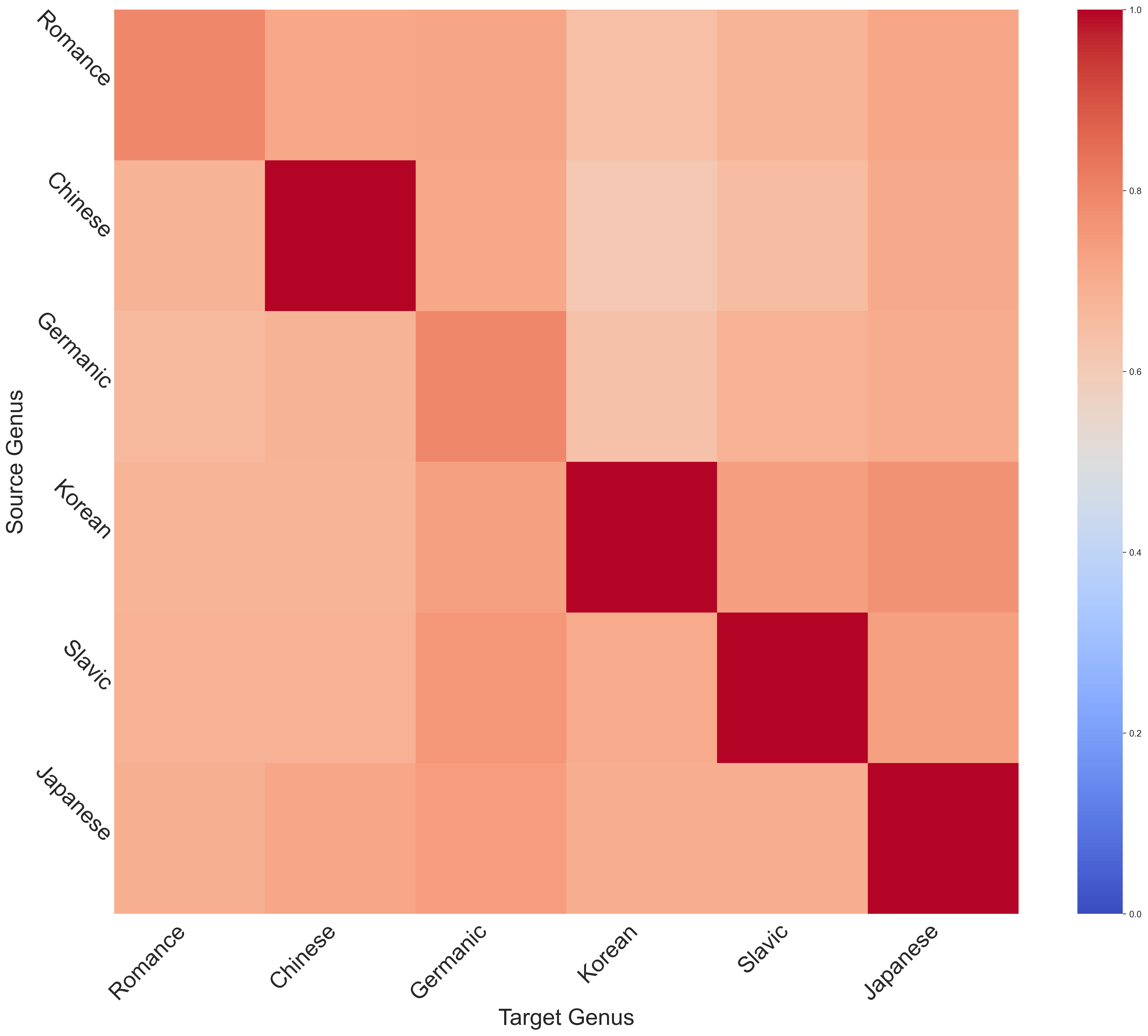}
        \caption{Switchscores of model Llama-7b, threshold 100.}
        \label{fig:switchscore_Llama7_th100}
    \end{subfigure}

    \caption{Switchscores Llama-7b.}
    \label{fig:switchscore_Llama7}
\end{figure}

\begin{figure}[!h]
    \centering
    \begin{subfigure}[b]{0.45\textwidth}
        \centering
        \includegraphics[width=0.7\textwidth]{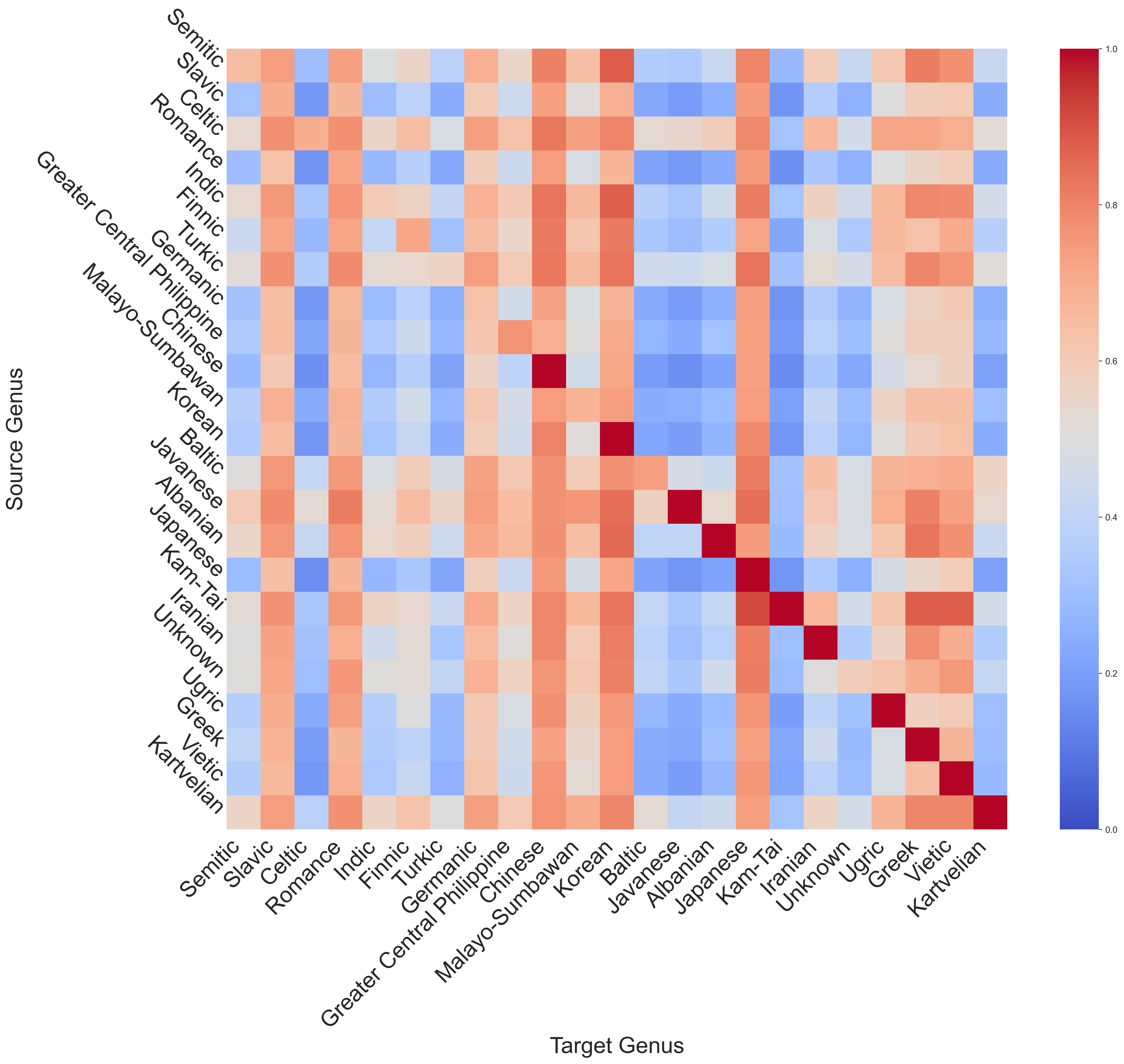}
        \caption{Switchscores of model Llama-13b, threshold 20.}
        \label{fig:switchscore_Llama13_th20}
    \end{subfigure}
    \hfill
    \begin{subfigure}[b]{0.45\textwidth}
        \centering
        \includegraphics[width=0.7\textwidth]{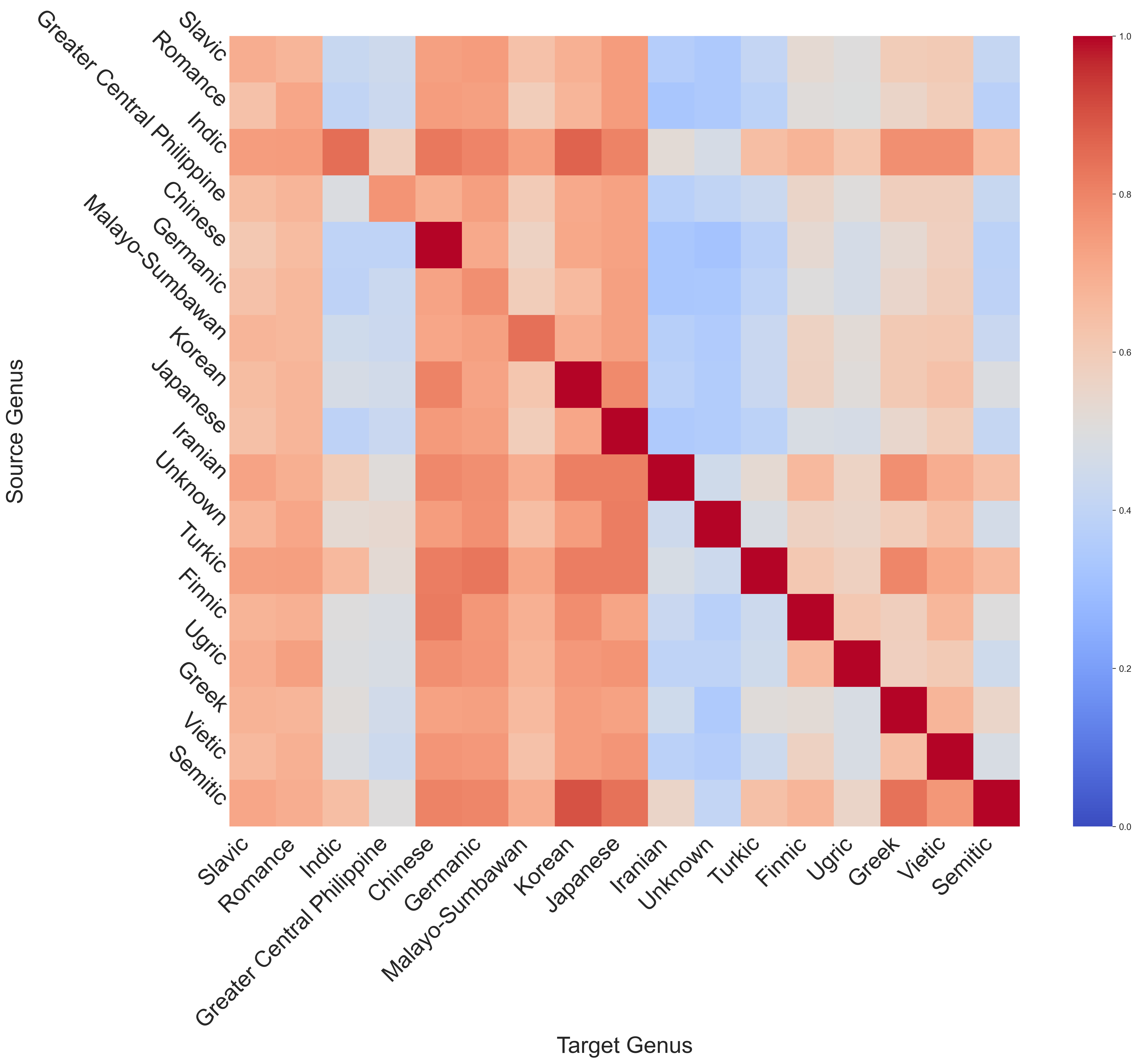}
        \caption{Switchscores of model Llama-13b, threshold 50.}
        \label{fig:switchscore_Llama13_th50}
    \end{subfigure}
    \hfill
    \begin{subfigure}[b]{0.45\textwidth}
        \centering
        \includegraphics[width=0.7\textwidth]{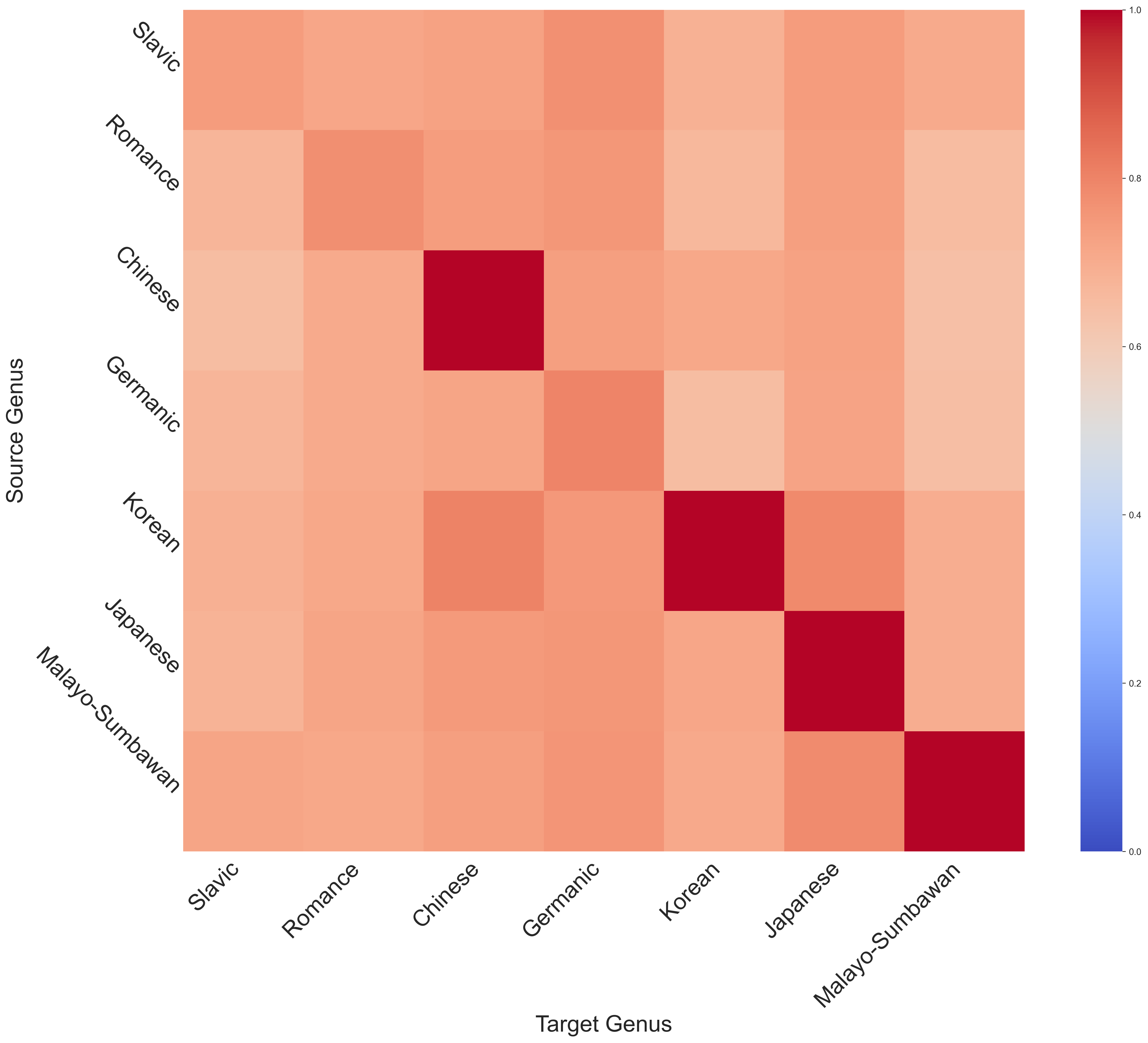}
        \caption{Switchscores of model Llama-13b, threshold 100.}
        \label{fig:switchscore_Llama13_th100}
    \end{subfigure}

    \caption{Switchscores Llama-13b.}
    \label{fig:switchscore_Llama13}
\end{figure}

\begin{figure}[!h]
    \centering
    \begin{subfigure}[b]{0.45\textwidth}
        \centering
        \includegraphics[width=0.7\textwidth]{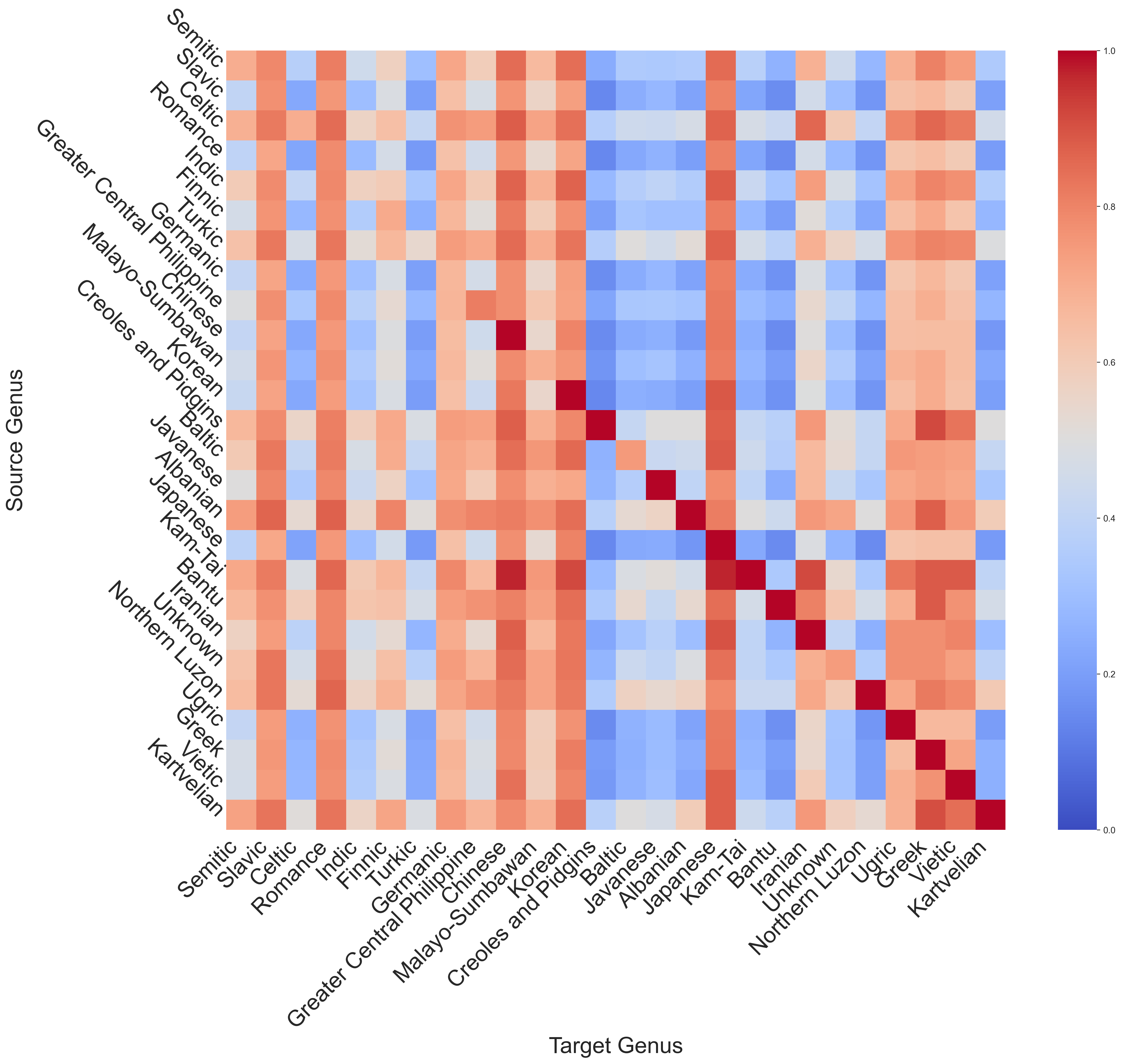}
        \caption{Switchscores of model Llama-70b, threshold 20.}
        \label{fig:switchscore_Llama70_th20}
    \end{subfigure}
    \hfill
    \begin{subfigure}[b]{0.45\textwidth}
        \centering
        \includegraphics[width=0.7\textwidth]{figs/Llama-70b/genus_th_50.png}
        \caption{Switchscores of model Llama-70b, threshold 50.}
        \label{fig:switchscore_Llama70_th50}
    \end{subfigure}
    \hfill
    \begin{subfigure}[b]{0.45\textwidth}
        \centering
        \includegraphics[width=0.7\textwidth]{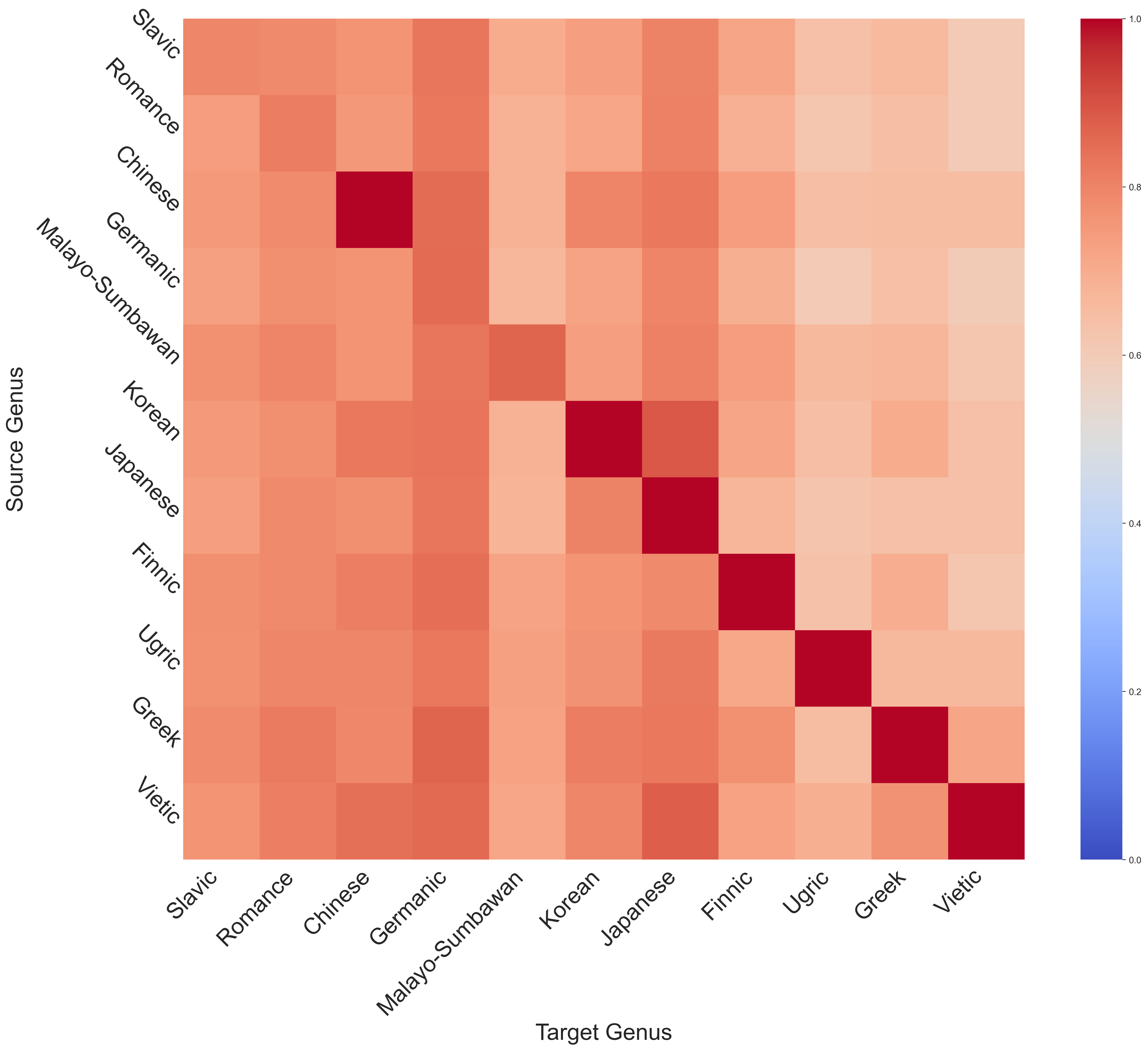}
        \caption{Switchscores of model Llama-70b, threshold 100.}
        \label{fig:switchscore_Llama70_th100}
    \end{subfigure}

    \caption{Switchscores Llama-70b.}
    \label{fig:switchscore_Llama70}
\end{figure}

\begin{figure}[!h]
    \centering
    \begin{subfigure}[b]{0.45\textwidth}
        \centering
        \includegraphics[width=0.7\textwidth]{figs/Mistral/genus_th_20.png}
        \caption{Switchscores of model Mistral-7B, threshold 20.}
        \label{fig:switchscore_Mistral_th20}
    \end{subfigure}
    \hfill
    \begin{subfigure}[b]{0.45\textwidth}
        \centering
        \includegraphics[width=0.7\textwidth]{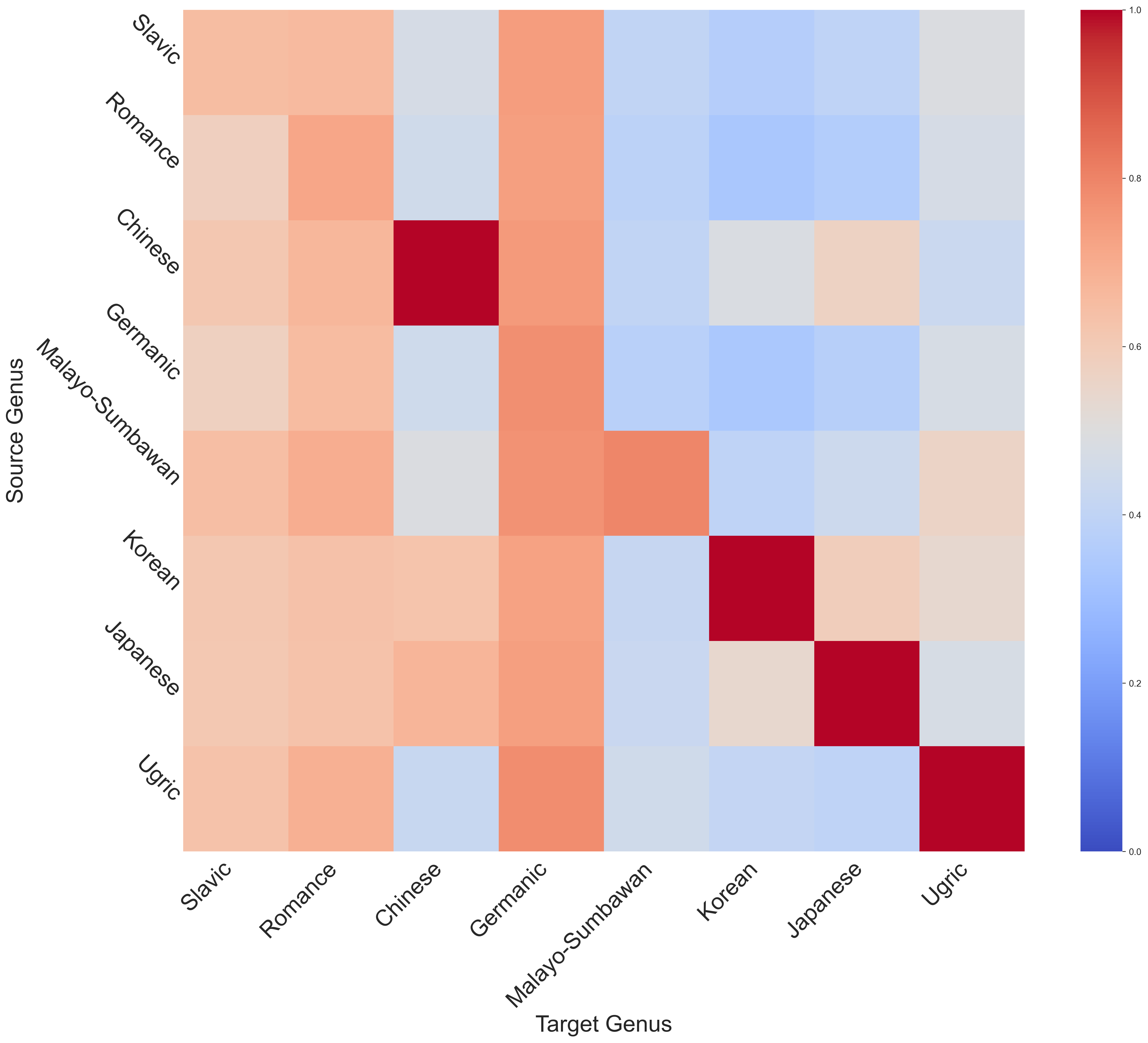}
        \caption{Switchscores of model Mistral-7B, threshold 50.}
        \label{fig:switchscore_Mistral_th50}
    \end{subfigure}
    \hfill
    \begin{subfigure}[b]{0.45\textwidth}
        \centering
        \includegraphics[width=0.7\textwidth]{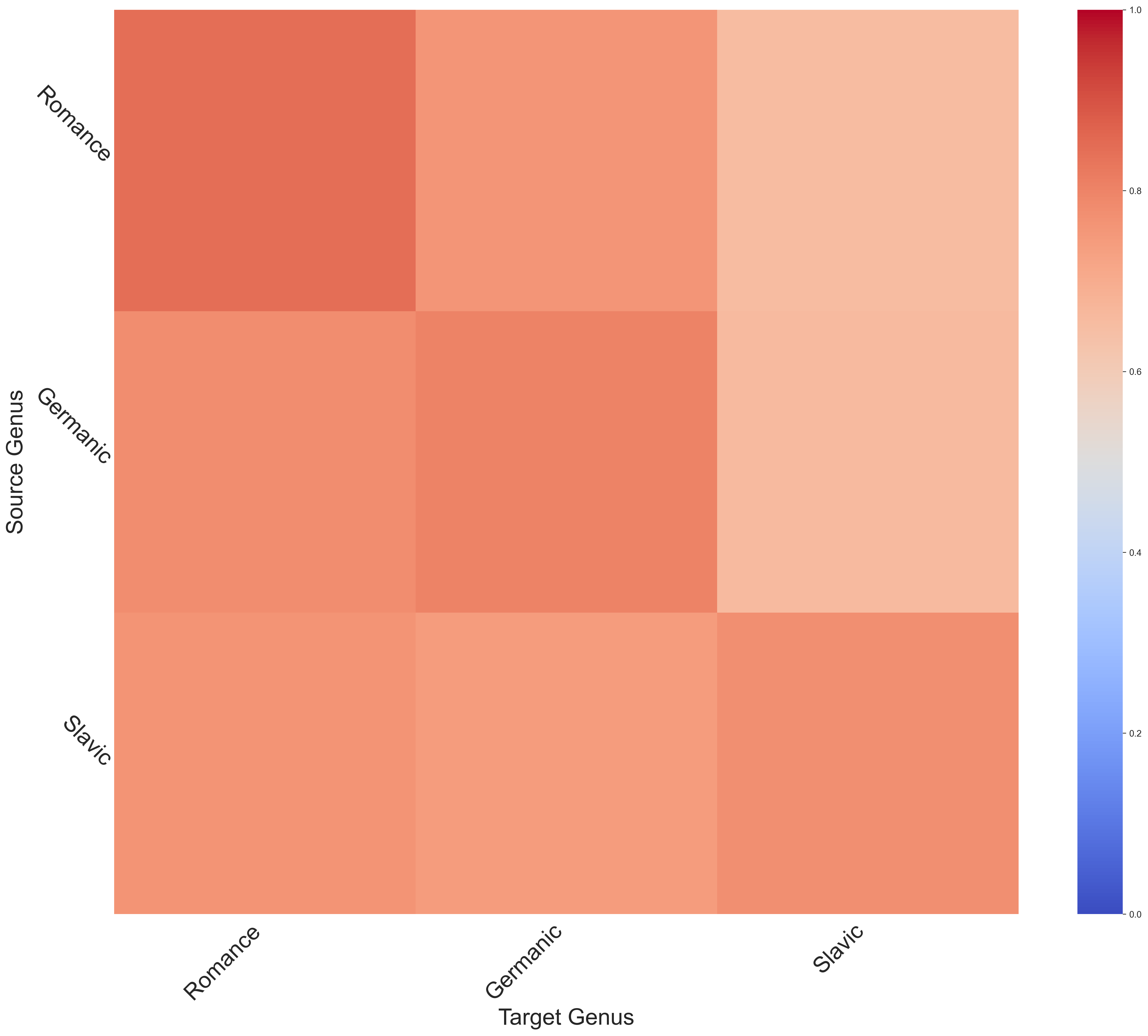}
        \caption{Switchscores of model Mistral-7B, threshold 100.}
        \label{fig:switchscore_Mistral_th100}
    \end{subfigure}

    \caption{Switchscores Mistral-7B.}
    \label{fig:switchscore_Mistral}
\end{figure}

\begin{figure}[!h]
    \centering
    \begin{subfigure}[b]{0.45\textwidth}
        \centering
        \includegraphics[width=0.7\textwidth]{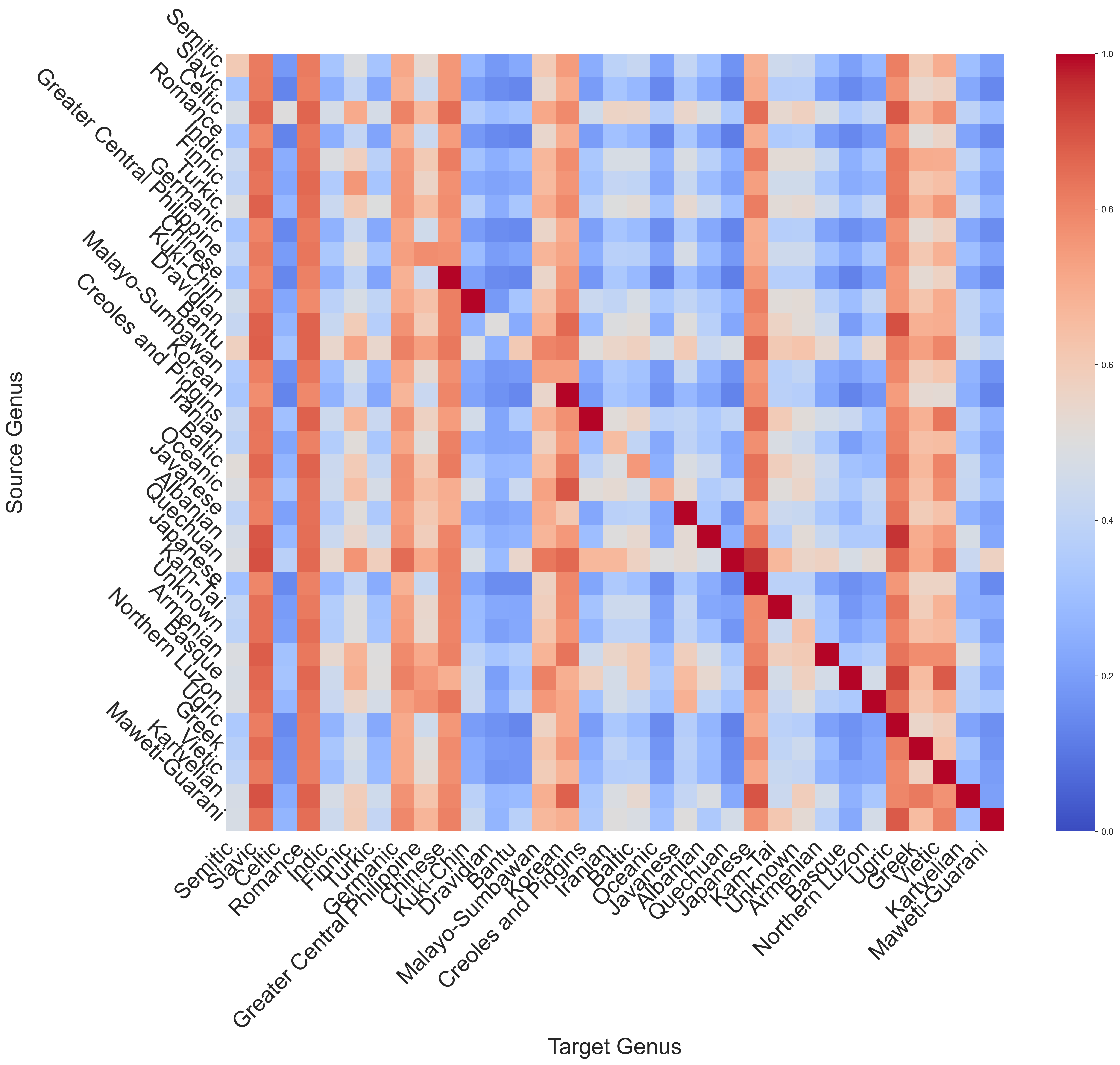}
        \caption{Switchscores of model Mixtral-8X7, threshold 20.}
        \label{fig:switchscore_Mixtral_th20}
    \end{subfigure}
    \hfill
    \begin{subfigure}[b]{0.45\textwidth}
        \centering
        \includegraphics[width=0.7\textwidth]{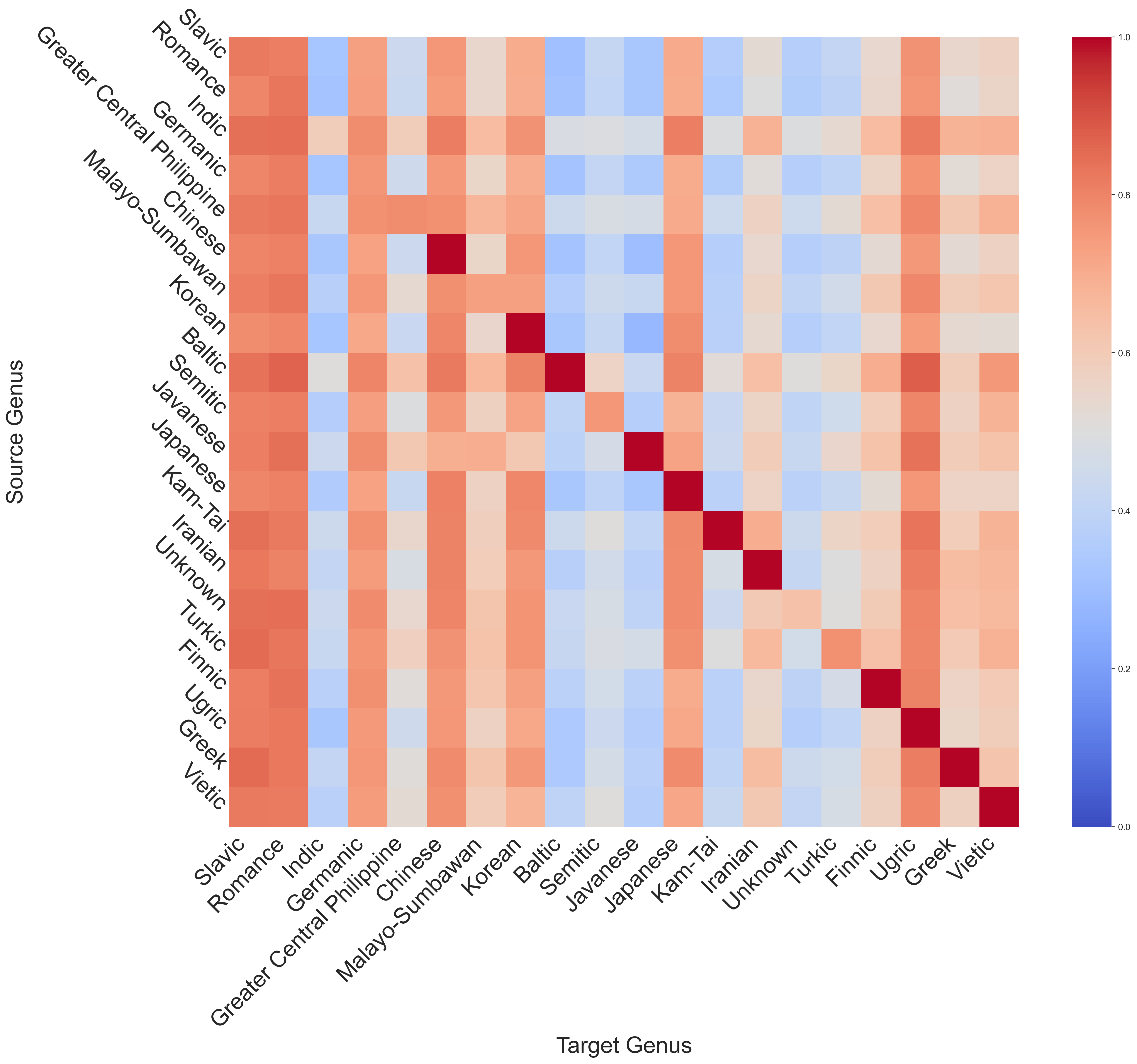}
        \caption{Switchscores of model Mixtral-8X7, threshold 50.}
        \label{fig:switchscore_Mixtral_th50}
    \end{subfigure}
    \hfill
    \begin{subfigure}[b]{0.45\textwidth}
        \centering
        \includegraphics[width=0.7\textwidth]{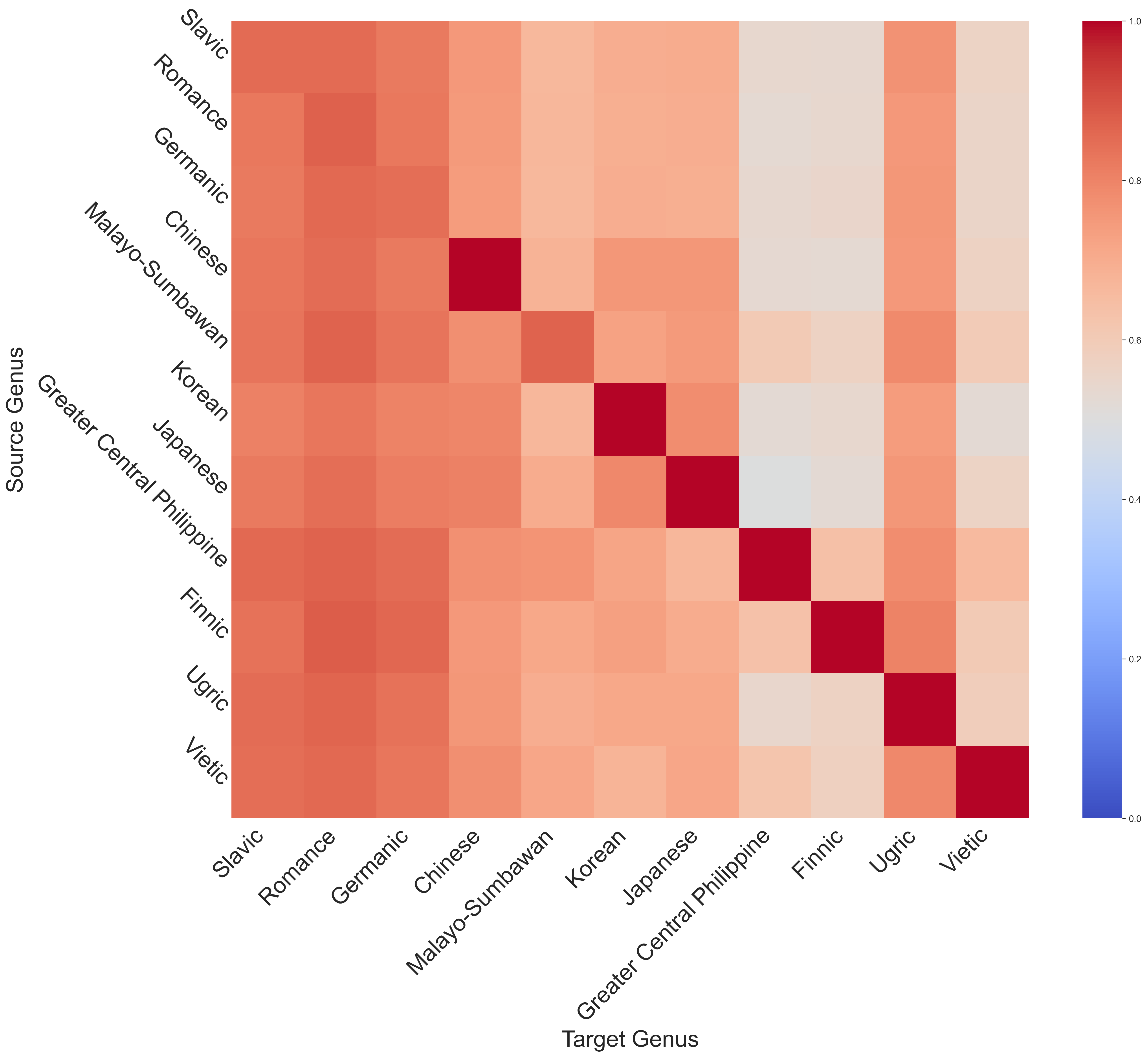}
        \caption{Switchscores of model Mixtral-8X7, threshold 100.}
        \label{fig:switchscore_Mixtral_th100}
    \end{subfigure}

    \caption{Switchscores Mixtral-8X7.}
    \label{fig:switchscore_Mixtral}
\end{figure}

\begin{figure}[!h]
    \centering
    \begin{subfigure}[b]{0.45\textwidth}
        \centering
        \includegraphics[width=0.7\textwidth]{figs/Qwen/genus_th_20.png}
        \caption{Switchscores of model Qwen-7B, threshold 20.}
        \label{fig:switchscore_Qwen_th20}
    \end{subfigure}
    \hfill
    \begin{subfigure}[b]{0.45\textwidth}
        \centering
        \includegraphics[width=0.7\textwidth]{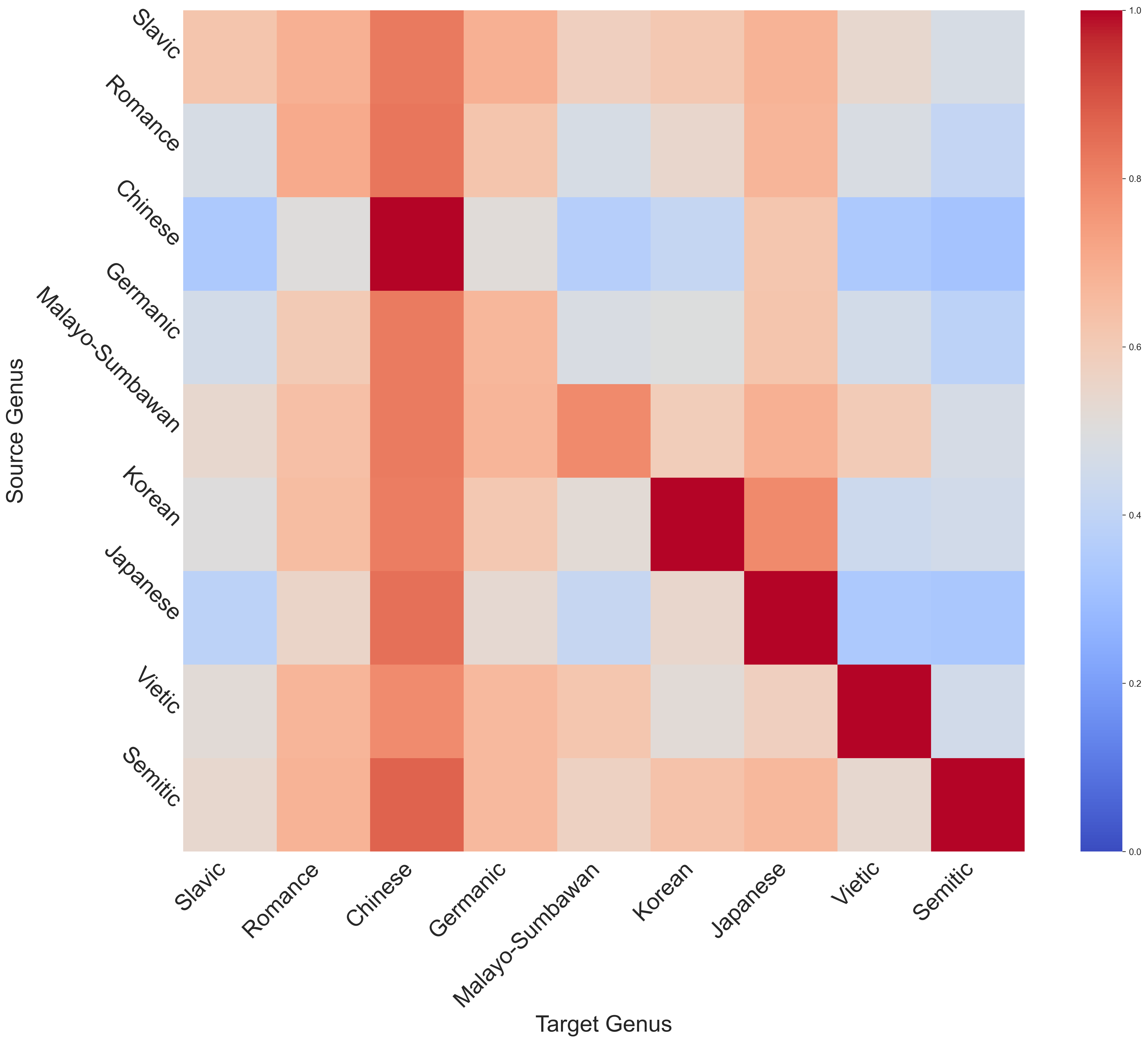}
        \caption{Switchscores of model Qwen-7B, threshold 50.}
        \label{fig:switchscore_Qwen_th50}
    \end{subfigure}
    \hfill
    \begin{subfigure}[b]{0.45\textwidth}
        \centering
        \includegraphics[width=0.7\textwidth]{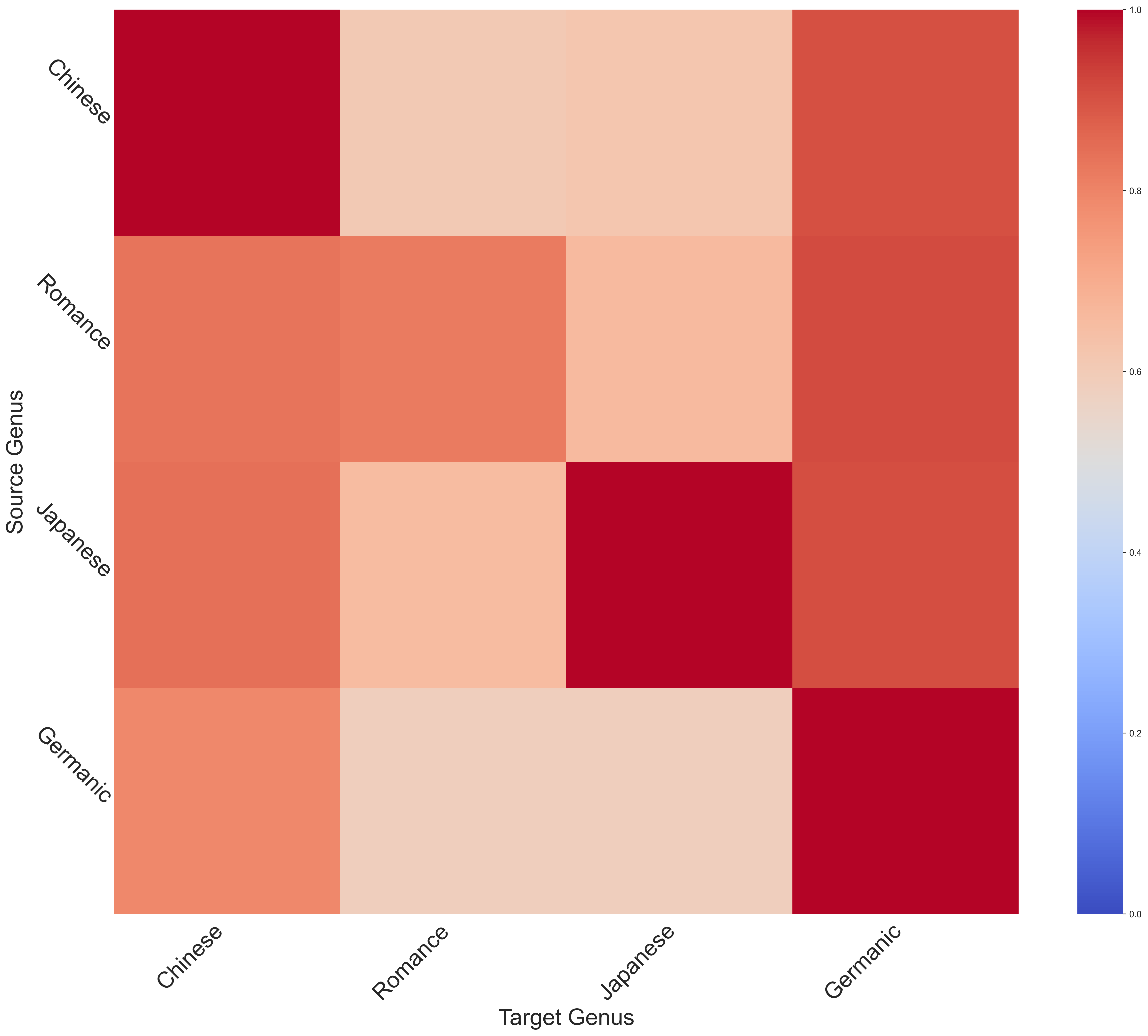}
        \caption{Switchscores of model Qwen-7B, threshold 100.}
        \label{fig:switchscore_Qwen_th100}
    \end{subfigure}

    \caption{Switchscores Qwen-7B.}
    \label{fig:switchscore_Qwen}
\end{figure}

\begin{figure}[!h]
    \centering
    \begin{subfigure}[b]{0.45\textwidth}
        \centering
        \includegraphics[width=0.7\textwidth]{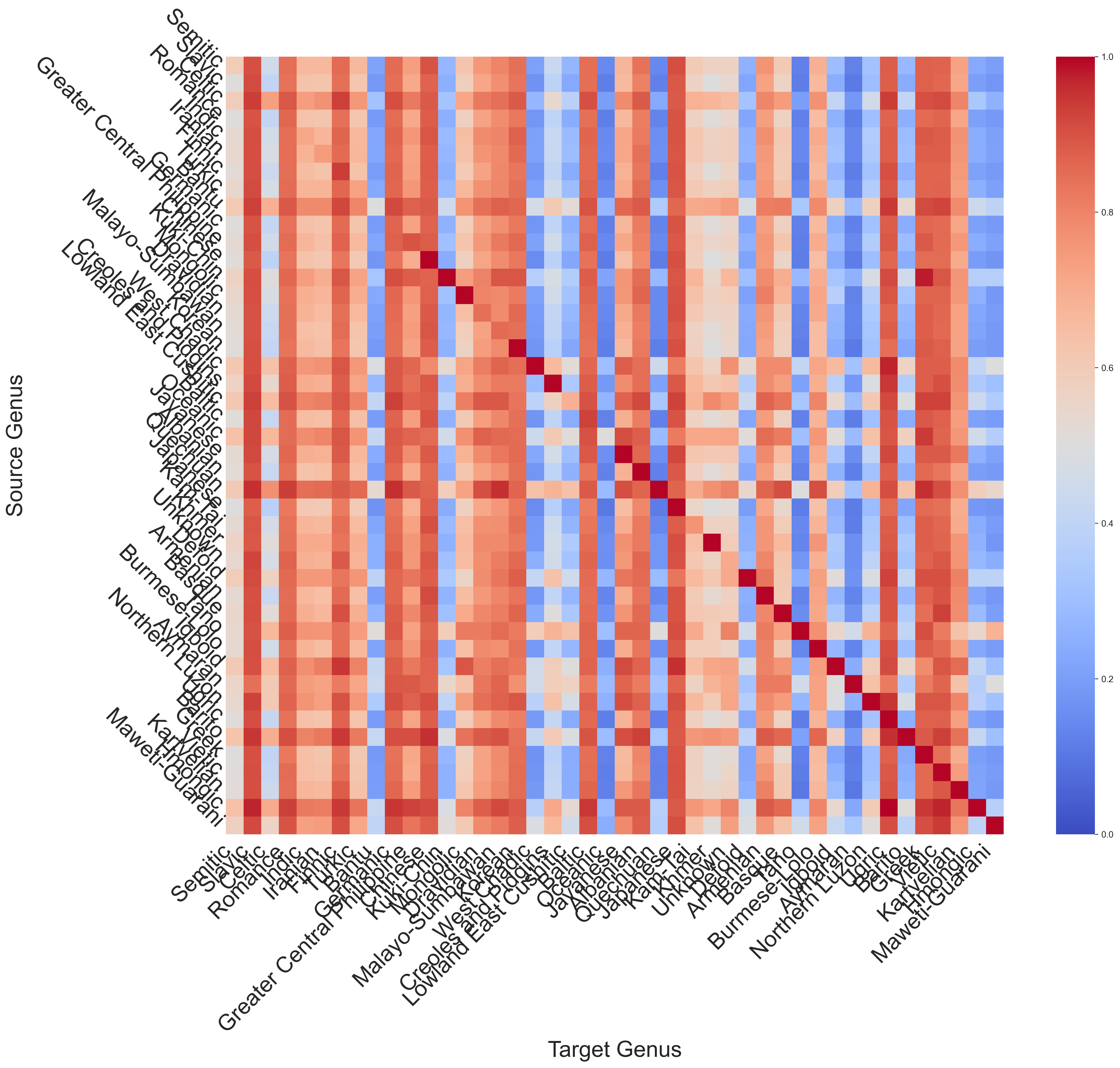}
        \caption{Switchscores of model Apertus-7B, threshold 20.}
        \label{fig:switchscore_Apertus_th20}
    \end{subfigure}
    \hfill
    \begin{subfigure}[b]{0.45\textwidth}
        \centering
        \includegraphics[width=\textwidth]{figs/Apertus/genus_th_50.png}
        \caption{Switchscores of model Apertus-7B, threshold 50.}
        \label{fig:switchscore_Apertus_th50}
    \end{subfigure}
    \hfill
    \begin{subfigure}[b]{0.45\textwidth}
        \centering
        \includegraphics[width=\textwidth]{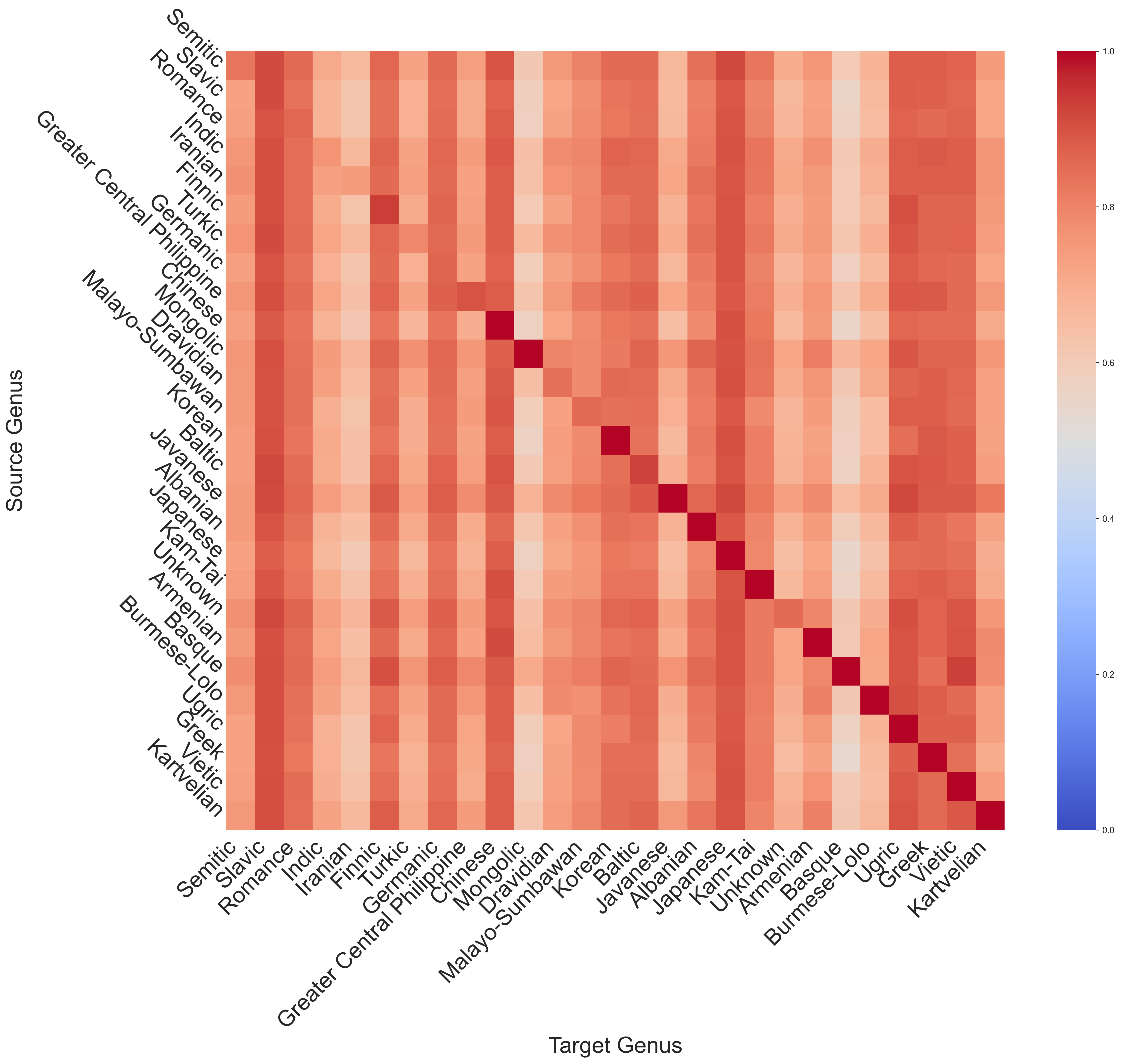}
        \caption{Switchscores of model Apertus-7B, threshold 100.}
        \label{fig:switchscore_Apertus_th100}
    \end{subfigure}

    \caption{Switchscores Apertus-7B.}
    \label{fig:switchscore_Apertus}
\end{figure}

\end{document}